\documentclass{article}

\usepackage[preprint]{neurips_2026}

\usepackage[utf8]{inputenc} 
\usepackage[T1]{fontenc}    
\usepackage{hyperref}       
\usepackage{url}            
\usepackage{amsmath}
\usepackage{booktabs}       
\usepackage{amsfonts}       
\usepackage{nicefrac}       
\usepackage{microtype}      
\usepackage{xcolor}         
\usepackage{multirow}
\usepackage{graphicx}
\hypersetup{  
    colorlinks=true,  
    citecolor=cyan,
}

\usepackage{amssymb}
\usepackage{fancyhdr}
\usepackage[most]{tcolorbox} 
\usepackage{tcolorbox} 
\usepackage[utf8]{inputenc}
\usepackage{graphicx}
\usepackage{tikz}
\usepackage[most]{tcolorbox}

\providecommand{\workshoptitle}[1]{}

\title{Learning to Foresee: Unveiling the Unlocking Efficiency of On-Policy Distillation}

\author{
  Yuchen Cai$^{1,2}$\thanks{These authors contributed equally to this work.}\thanks{This work was done during an internship at Tencent.},
  Ding Cao$^{1,4}$\footnotemark[1],
  Liang Lin$^{5}$,
  Chunxi Luo$^{6}$,
  Xin Xu$^{2}$,
  Kai Yang$^{2}$,
  Weijie Liu$^{2}$, \\
  \bfseries Saiyong Yang$^{2}$,
  Tianxiang Zhao$^{4}$,
  Guangzhong Sun$^{1}$,
  Guiquan Liu$^{1}$\thanks{Corresponding authors: \texttt{gqliu@ustc.edu.cn}, \texttt{fjf@mail.ustc.edu.cn}},
  Junfeng Fang$^{3}$\footnotemark[3]
  \\
  \\
  $^{1}$USTC,
  $^{2}$Tencent,
  $^{3}$NUS,
  $^{4}$HKUST(GZ),
  $^{5}$UCAS-IIE,
  $^{6}$SHU
  \\
  \\
  \texttt{\{caiyuchen,caoding\}@mail.ustc.edu.cn}
}

\begin{document}
\noindent
\includegraphics[height=0.6cm]{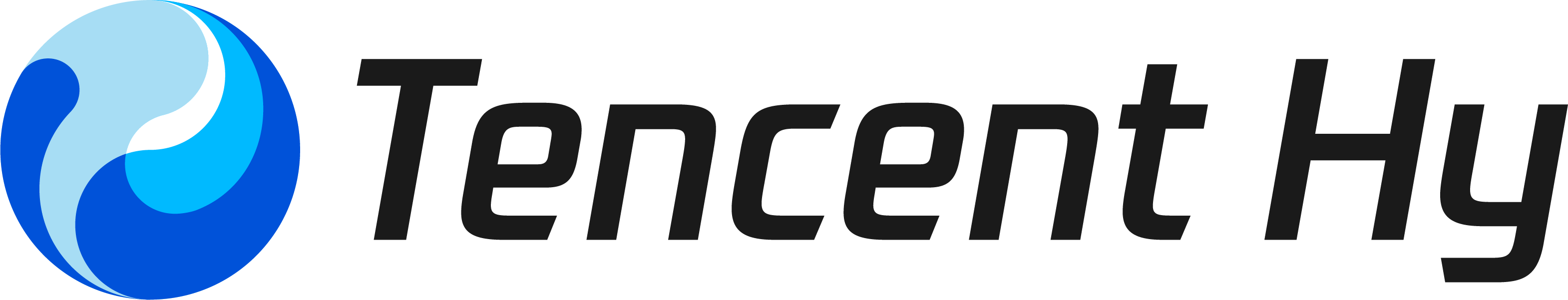}
\hfill   
\textcolor{gray}{\fontsize{11pt}{10pt}\selectfont May 13, 2026}
\vskip -0.1in
\maketitle


\begin{abstract}

On-policy distillation (OPD) has emerged as an efficient post-training paradigm for large language models. However, existing studies largely attribute this advantage to denser and more stable supervision, while the parameter-level mechanisms underlying OPD's efficiency remain poorly understood. In this work, we argue that OPD's efficiency stems from a form of ``foresight'': it establishes a stable update trajectory toward the final model early in training. This foresight manifests in two aspects. First, at the \textbf{Module-Allocation Level}, OPD identifies regions with low marginal utility and concentrates updates on modules that are more critical to reasoning. Second, at the \textbf{Update-Direction Level}, OPD exhibits stronger low-rank concentration, with its dominant subspaces aligning closely with the final update subspace early in training. Building on these findings, we propose \textbf{EffOPD}, a plug-and-play acceleration method that speeds up OPD by adaptively selecting an extrapolation step size and moving along the current update direction. EffOPD requires no additional trainable modules or complex hyperparameter tuning, and achieves an average training acceleration of $3\times$ while maintaining comparable final performance. Overall, our findings provide a parameter-dynamics perspective for understanding the efficiency of OPD and offer practical insights for designing more efficient post-training methods for large language models. Our code is available at: \href{https://github.com/caiyuchen-ustc/EffOPD}{https://github.com/caiyuchen-ustc/EffOPD}.
\end{abstract}

\begin{figure}[t]
    \includegraphics[width=1\textwidth]{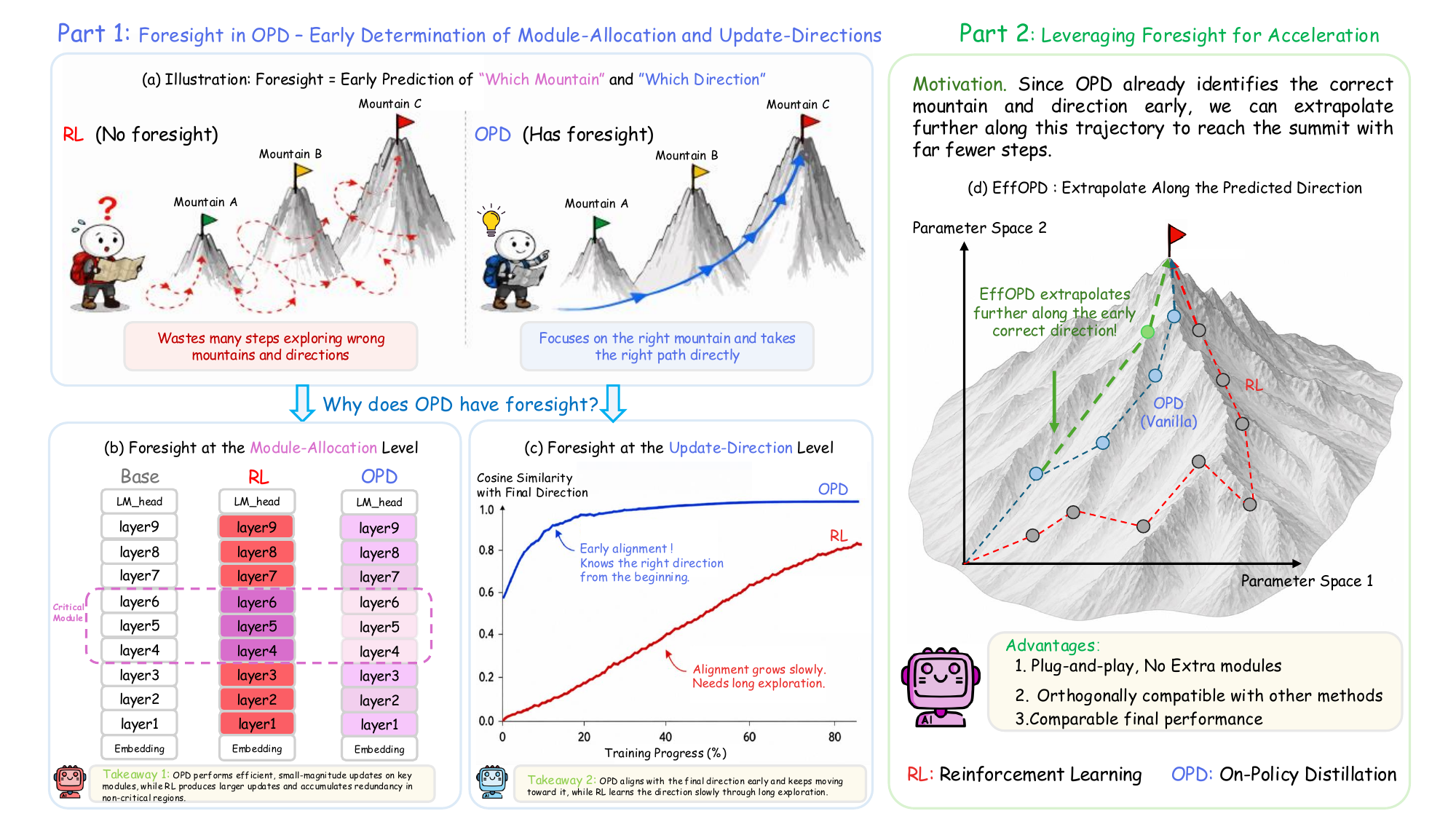}
    \caption{Illustration of the foresight mechanism in OPD. Compared with RL, OPD identifies critical modules and aligns with the final optimization direction early in training, concentrating effective updates while reducing redundancy. Based on this, we propose EffOPD, which extrapolates along the early predicted direction to accelerate training.}
    \label{introfig}
    \vskip -0.2in
\end{figure}

\begin{flushright}
\small\itshape
``To foresee the future is to master the present.''\\
--- Niccolò Machiavelli
\end{flushright}

\section{Introduction}


As large language models (LLMs) continue to advance in reasoning \citep{OpenAI2025,deepseekai2025deepseekr1incentivizingreasoningcapability}, On-Policy Distillation (OPD) has emerged as an important paradigm for post-training and model fusion \citep{agarwal2024onpolicydistillationlanguagemodels,xiao2026mimo,deepseek2026v4}. Given a teacher model, OPD leverages dense supervisory signals to achieve performance comparable to Reinforcement Learning (RL) with substantially reduced training time \citep{Venkatkrishna2026AletheiaWM,yang2025qwen3technicalreport}. Existing studies mainly attribute this advantage to denser and more stable supervision \citep{he2026far,yue2025doesreinforcementlearningreally}. However, such optimization-centric explanations remain largely macroscopic and fail to capture the underlying parameter update dynamics \citep{zhang2025surveyreinforcementlearninglarge}.

In this work, we argue that OPD's efficiency stems from a form of ``foresight'': it establishes stable and highly aligned update directions early in training, enabling rapid convergence with limited exploration and correction. This foresight manifests in two aspects.

\textbf{Foresight at the Module-Allocation Level.} 
Our analysis reveals that, under the same update norm constraint, OPD achieves larger performance gains than RL, suggesting that its advantage does not merely stem from the magnitude of parameter updates \citep{geva2021transformerfeedforwardlayerskeyvalue,geva2023dissectingrecallfactualassociations}. Further analysis shows that, although RL and OPD exhibit similar sensitivity patterns across layers and modules, RL accumulates substantially larger update norms in modules with limited contribution to performance improvement, thereby introducing redundant updates with low marginal utility. In contrast, OPD demonstrates a form of ``foresight''. As shown in Figure~\ref{introfig} (b), it identifies these low-utility modules early in training and suppresses their parameter updates, allowing updates to concentrate more effectively on intermediate-layer modules that are more critical to reasoning \citep{meng2023locatingeditingfactualassociations}.

\textbf{Foresight at the Update-Direction Level.}
At the update-direction level, OPD's foresight lies in the early alignment between its update directions and the principal directions of the final solution. Spectral and subspace evolution analyses show that OPD concentrates updates on a few stable dominant directions early in training \citep{zhang2015singularvaluedecompositionapplications}, whose dominant directions are highly aligned with the final update subspace and remain stable thereafter, as shown in Figure~\ref{introfig} (c). In contrast, RL exhibits more dispersed updates, with delayed and more fluctuating alignment. Moreover, after module-wise norm scaling, an OPD checkpoint at only 10\% training progress recovers approximately 80\% of the final reasoning performance. This suggests that OPD captures the main structure of the final solution early and locks onto an effective direction with minimal exploration and correction.

To further validate these insights and improve the training efficiency of OPD, we propose \textbf{EffOPD}, a simple and intuitive acceleration framework. As shown in Figure~\ref{introfig} (d), EffOPD performs linear extrapolation along the current update direction, leveraging the inherent ``foresight'' of OPD to match the final performance of vanilla OPD with fewer training iterations and samples. Experiments across model scales from 1.5B to 32B parameters show that EffOPD achieves an average training acceleration of $3\times$ over multiple baselines in a plug-and-play manner, while maintaining comparable final performance.

In summary, this work identifies a form of foresight in OPD for LLMs and argues that it is a key source of its training efficiency. Our analysis provides a parameter-level explanation for the common intuition that distillation is easier to optimize due to denser supervision \citep{Yang2026LearningBT}. Building on these findings, EffOPD offers a simple plug-and-play acceleration method for OPD, requiring no additional modules, complex hyperparameter tuning, or human intervention. It achieves an average training acceleration of $3\times$ and remains orthogonal to existing acceleration techniques, providing new insights into the design of more interpretable and efficient post-training paradigms for large language models.

\begin{figure}[t]
    \includegraphics[width=1\textwidth]{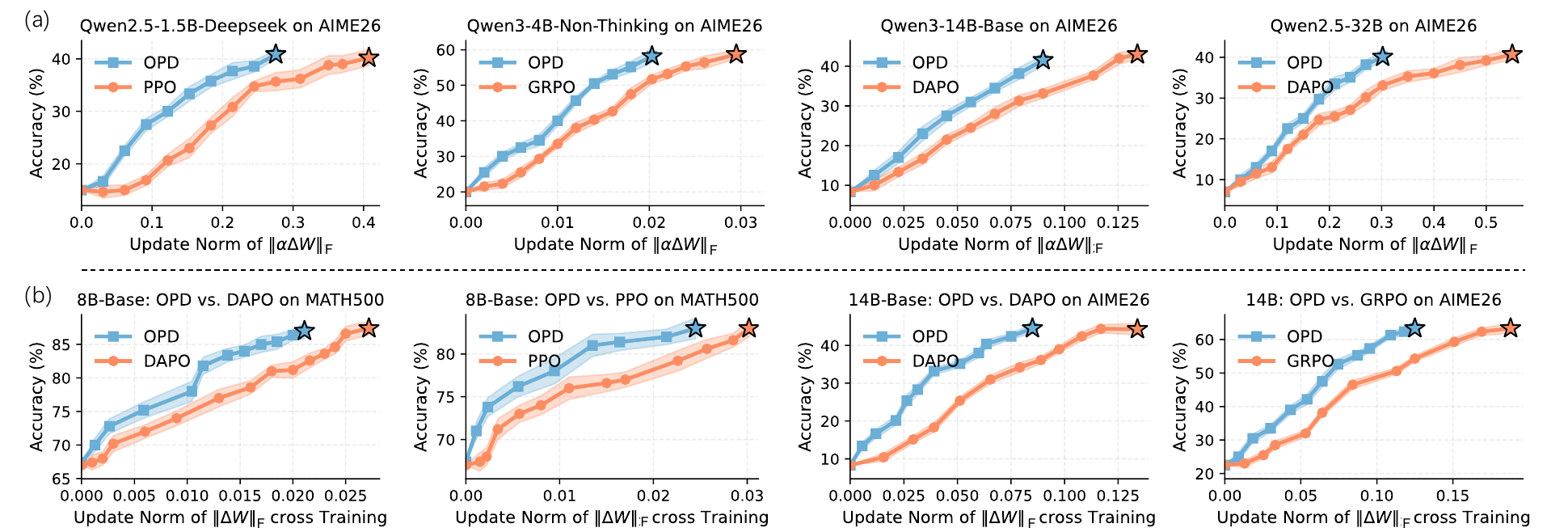}
    \caption{Comparison of parameter update efficiency between RL and OPD. (a) Scaling analysis at the final checkpoint: for updates scaled to the same norm, OPD achieves substantially higher reasoning gains than RL. (b) Training dynamics: across the entire optimization trajectory, OPD consistently requires smaller parameter updates than RL to reach equivalent reasoning accuracy.}
    \label{fig1}
\end{figure}

\section{Functional Redundancy Avoidance}
\label{section2}
In this section, we investigate the modular-level differences between OPD and RL. We show that OPD exhibits modular-level ``foresight'': it preferentially concentrates updates in high-marginal-utility functional regions while suppressing parameter changes in low-utility regions. We refer to this property as \textbf{Functional Redundancy Avoidance}. Section~\ref{Experimental Setting} introduces the experimental setup, and Section~\ref{Parameter Updates and Reasoning Gains} compares OPD with RL to show how this foresight leads to more compact and efficient parameter updates.

\subsection{Experimental Setting}
\label{Experimental Setting}
Our analysis uses a shared initialization $W_{\mathrm{Base}}$ for both RL and OPD, with parameter updates defined as $\Delta W_{\mathrm{RL/OPD}} = W_{\mathrm{RL/OPD}} - W_{\mathrm{Base}}$. We conduct experiments across models ranging from 1.5B to 32B parameters, including pretrained, SFT-tuned, and Thinking-series models~\citep{qwen2025qwen25technicalreport,zhang2025instructiontuninglargelanguage,yang2025qwen3technicalreport}. For RL, we consider PPO, GRPO, and DAPO~\citep{yu2025dapoopensourcellmreinforcement}. For OPD, the student is trained with a pattern-aligned teacher, typically a stronger model from the same family~\citep{li2026rethinking}. Further details are provided in Appendix~\ref{Experimental Setup}.

\subsection{Parameter Updates \& Reasoning Gains}
\label{Parameter Updates and Reasoning Gains}

\paragraph{Results on Fully Trained Models.}
We first examine the update efficiency at the final checkpoint. Specifically, we fix the update direction $\Delta W_{\mathrm{RL/OPD}}$ from the last checkpoint and scale its magnitude using a factor $\alpha \in [0,1]$, evaluating models of the form $W_{\mathrm{Base}} + \alpha \Delta W_{\mathrm{RL/OPD}}$. As shown in Figure~\ref{fig1} (a), when updates are scaled to the same norm, OPD achieves substantially higher reasoning gains than RL. This indicates that $\Delta W_{\mathrm{RL}}$ contains a non-negligible number of components weakly correlated with task performance—they contribute to the update norm but provide limited reasoning improvement. In contrast, OPD updates carry a greater fraction of task-relevant signal that effectively translates into performance gains.

\paragraph{Results across the Training Process.}
This observation naturally raises a key question: when do these weakly task-correlated components emerge during RL training? Since the performance of RL-trained models typically saturates in later stages, one possible explanation is that redundant updates mainly accumulate near the end of training~\citep{khatri2025art,zheng2025stabilizingreinforcementlearningllms}. To examine this, we analyze intermediate checkpoints of both RL and OPD throughout training and track the relationship between parameter update magnitude and reasoning accuracy. As shown in Figure~\ref{fig1} (b), OPD consistently requires smaller parameter updates than RL to achieve the same reasoning accuracy. Moreover, OPD achieves rapid accuracy improvement with relatively small increases in $\Delta W_{\mathrm{OPD}}$ norm, whereas RL improves more slowly under comparable update magnitudes. These results suggest that OPD's superior efficiency does not simply come from avoiding late-stage redundancy, but from forming a compact and task-relevant update pattern early in training.

\begin{figure}[t]
    \includegraphics[width=1\textwidth]{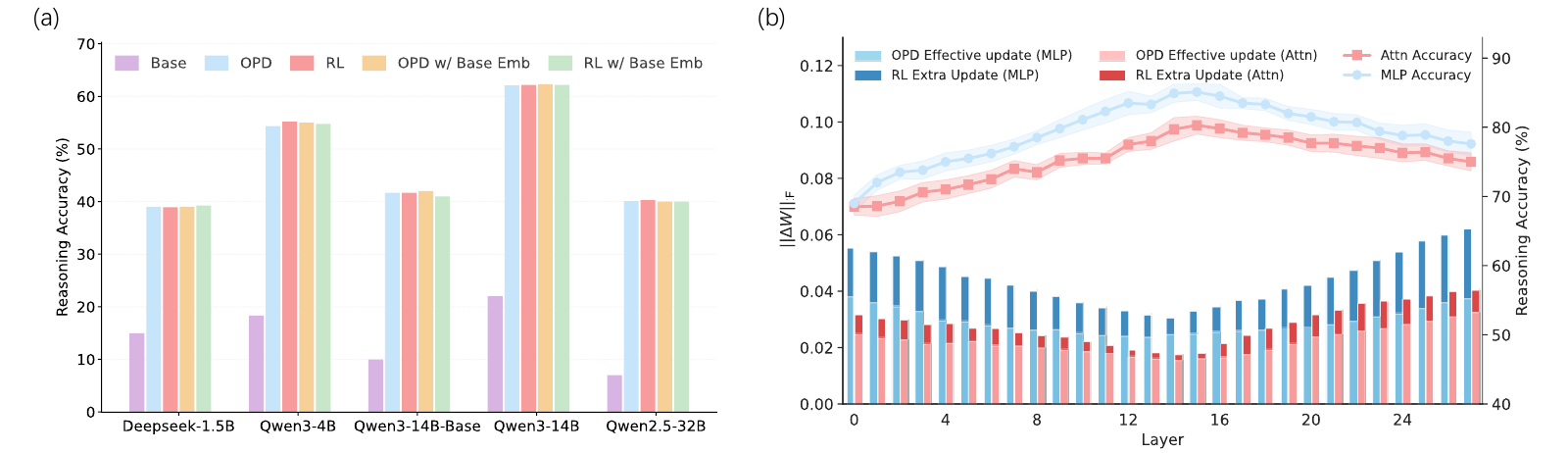}
    \caption{Functional contributions and update distributions across architectural components. 
(a) Effect of embedding layer replacement on AIME26. 
(b) Layer-wise update norms (bars, left axis) for RL/OPD-trained Qwen3-8B-Base models, and corresponding OPD reasoning accuracy after sliding-window intervention (line, right axis) on MATH500.}
    \label{fig2}
\end{figure}

\paragraph{Locating the Redundant Updates.}
\label{Locating the Residual Updates}

The previous analysis shows that RL updates contain components with relatively low task relevance. To locate these redundancies and assess their functional contributions, we decompose model updates into three architectural components: embedding, MLP, and attention layers. We first examine the embedding layer by replacing the embeddings of OPD and RL models with those from the base model while keeping all other parameters unchanged. As shown in Figure~\ref{fig2} (a), this intervention has negligible impact on reasoning performance, suggesting that embedding updates contribute little to reasoning gains. Thus, the main functional updates of OPD and RL are likely concentrated in deeper model components rather than the embedding layer.

Next, we conduct a sliding-window intervention analysis to locate the functional regions of OPD and RL updates. Following prior block-wise intervention studies~\citep{cai2024locatingmitigatinggenderbias, meng2023locatingeditingfactualassociations}, we partition the model into consecutive layer blocks and inject local OPD or RL updates into each block to evaluate their impact on reasoning performance\footnote{Detailed setup is provided in Appendix~\ref{Detailed Setup of Sliding-Window Intervention Analysis}.}. As shown in Figure~\ref{fig2} (b) and Figure~\ref{appendix3} (b), MLP modules are overall more sensitive to reasoning-related updates than attention modules, indicating that MLPs serve as the primary carriers of knowledge representation and relational reasoning. From the perspective of layer position, the performance curves of both module types exhibit a clear inverted U-shaped pattern: interventions in the middle layers yield the largest gains, whereas those in the bottom and top layers lead to relatively smaller improvements. This suggests that reasoning-related updates are not uniformly distributed across the network, but are mainly concentrated in middle-layer MLPs with stronger functional coupling. These findings are consistent with prior mechanistic interpretability studies on the functional roles of Transformer modules and layers~\citep{skean2025layer, geva2021transformerfeedforwardlayerskeyvalue, geva2022transformerfeedforwardlayersbuild}.

Building on these observations, we further compare the update patterns of OPD and RL. The two methods exhibit highly consistent intervention sensitivity distributions across both module types and layer positions, suggesting that OPD and RL do not rely on fundamentally different functional pathways, but instead optimize along the model's existing key functional structures. The key difference lies in their layer-wise update norms. RL introduces substantially larger parameter changes in the low-sensitivity bottom and top layers. Since interventions in these peripheral layers yield limited performance gains, their larger update norms do not translate into proportional performance gains and are therefore more likely to reflect redundant updates weakly related to task rewards. In contrast, while maintaining a functional update distribution similar to RL, OPD significantly suppresses parameter changes in low-sensitivity regions and concentrates updates more strongly in middle-layer modules with higher functional contributions. Therefore, the advantage of OPD does not come from learning an entirely new update mechanism, but from more accurately distinguishing high-benefit from low-benefit parameter regions and reducing ineffective updates in peripheral layers, thereby achieving higher update efficiency and stronger reasoning performance gains with more compact parameter changes. Additionally, we further present the visualized differences and performance comparison results between RL and OPD across different components. We recommend interested readers to refer to the detailed results and analysis in Appendix~\ref{Property 1 Additional Experiment}.

\paragraph{Summary.}
The above results show that OPD exhibits clear foresight at the modular level, which we formalize as Property~1: \textbf{Functional Redundancy Avoidance}. Compared with RL, OPD forms a compact and task-relevant update pattern earlier in training, suppresses redundant parameter changes in low-marginal-utility regions, and concentrates updates in reasoning-critical modules with higher functional contributions, thereby achieving higher update efficiency and stronger reasoning performance gains.



\begin{table}[t]
\centering
\caption{Characterization of Parameter Update Geometry: OPD vs.\ RL Across Model Scales.}
\label{table1}
\setlength{\tabcolsep}{2.5pt}
\renewcommand{\arraystretch}{0.95}
\begin{tabular}{l *{4}{cc}}
\toprule
\multirow{2}{*}{Metric}
& \multicolumn{2}{c}{\textbf{1.5B}}
& \multicolumn{2}{c}{\textbf{4B}}
& \multicolumn{2}{c}{\textbf{8B}}
& \multicolumn{2}{c}{\textbf{14B}} \\
\cmidrule(lr){2-3}\cmidrule(lr){4-5}\cmidrule(lr){6-7}\cmidrule(lr){8-9}
& \textsc{rl} & \textsc{opd}
& \textsc{rl} & \textsc{opd}
& \textsc{rl} & \textsc{opd}
& \textsc{rl} & \textsc{opd} \\
\midrule
Spectral Norm ($\uparrow$)
& 0.094 & 0.113
& 0.007 & 0.009
& 0.004 & 0.005
& 0.056 & 0.063 \\
Spectral / Frobenius Norm Ratio ($\uparrow$)
& 33.2\% & 39.6\%
& 19.7\% & 25.7\%
& 32.7\% & 36.8\%
& 24.4\% & 28.1\% \\
Effective Rank ($\downarrow$)
& 964  & 778
& 1908 & 1587
& 2754 & 2341
& 3174 & 2937 \\
Top-1\% Subspace Norm Ratio ($\uparrow$)
& 78.1\% & 92.3\%
& 79.2\% & 93.4\%
& 88.5\% & 94.7\%
& 81.2\% & 94.5\% \\
\bottomrule
\end{tabular}
\end{table}

\section{Early Low-Rank Lock-in}
\label{section3}

The preceding analysis reveals OPD's ``foresight'' at the modular level. Building on this, we further investigate the intrinsic organization of its parameter updates from a geometric perspective and introduce the property \textbf{Early Low-Rank Lock-in} to describe this potential structural constraint. Specifically, we validate this property by analyzing the spectral concentration of the update matrix, the functional contributions of different subspaces, and the functional effectiveness of early stabilized directions through norm scaling experiments.

\subsection{Spectral Concentration of Update Matrix}
\label{Spectral Concentration of Update Matrix}
To characterize the spectral structure of parameter updates, we perform singular value decomposition (SVD)~\citep{5197422} on the update matrix $\Delta W_{\mathrm{RL/OPD}} = U\Sigma V^\top$ and introduce four complementary geometric metrics\footnote{Detailed definitions are provided in Appendix~\ref{Geometric Metrics for Parameter Update Matrix}.}: \textbf{Spectral Norm}~\citep{MATHIAS1990269}, \textbf{Spectral / Frobenius Norm Ratio}~\citep{ALNATOOR2024e30056}, \textbf{Effective Rank}~\citep{item_f3c74b8f1cad43ed869604b318d58703}, and \textbf{Top-1\% Subspace Norm Ratio}~\citep{cai2025predictability}. The first two metrics quantify the dominance of leading singular directions, while the latter two measure the concentration of update energy across the spectrum. Table~\ref{table1} reports the average values over all MLP and attention matrices. Across all model scales, OPD consistently exhibits stronger low-rank structure than RL. For example, on the 8B model, OPD achieves a higher spectral-to-Frobenius norm ratio (36.8\% vs.\ 32.7\%), lower effective rank (2341 vs.\ 2754), and higher Top-1\% subspace norm ratio (94.7\% vs.\ 88.5\%). These results suggest that OPD concentrates update energy into a small set of dominant directions more effectively than RL. Notably, despite having a smaller overall update norm, OPD allocates a larger proportion of its update energy to these dominant subspaces. This raises a key question: does such directional concentration explain the efficiency advantage of OPD observed in Section~\ref{section2}? To answer this, we conduct two controlled experiments to separately examine the roles of update direction and update magnitude.

\begin{figure}[t]
    \includegraphics[width=1\textwidth]{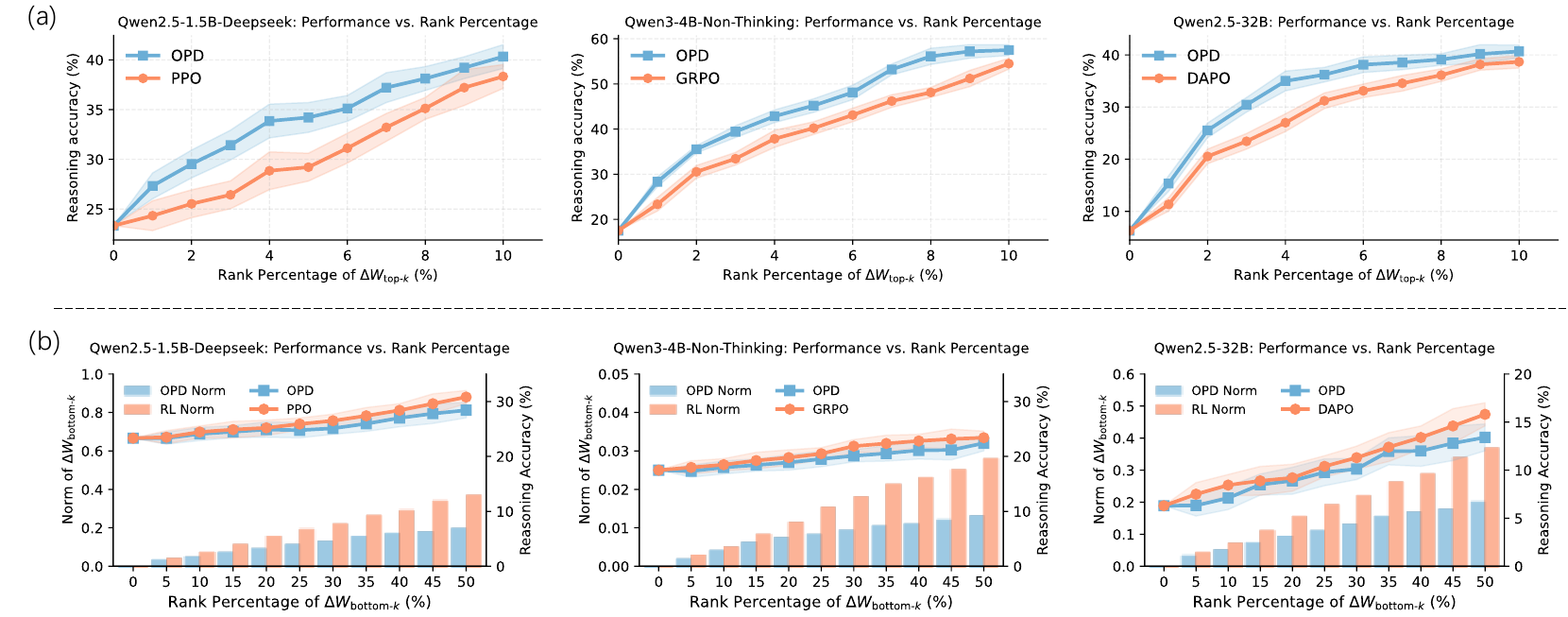}
    \caption{Low-rank subspace analysis. (a) Top-$k\%$ subspace: OPD achieves higher performance; (b) Bottom-$k\%$ subspace: RL incurs significantly larger norm cost for marginal performance gains.}
    \label{fig3}
\end{figure}

\subsection{Functional Partition of the Update Spectrum: Principal vs. Tail Subspaces}
\label{Functional Partition of the Update Spectrum: Principal vs. Tail Subspaces}
\paragraph{Top-$k\%$ Subspace: Directional Quality under Equal Norm Budget.}

To assess the intrinsic directional quality of the principal subspace, we construct a Top-$k\%$ truncated approximation $\Delta W_{\text{Top-}k\%}$ using the Top-$k\%$ singular components, and subsequently rescale its Frobenius norm to match between RL and OPD. After applying this low-rank update to the base model, we evaluate its reasoning performance. By standardizing the norm budget, we are able to directly compare the directional quality of the Top-$k\%$ principal subspaces between RL and OPD.

As shown in Figure~\ref{fig3} (a), both methods recover over 95\% of their full-model reasoning performance using only 10\% of the rank, confirming that the Top-$k\%$ subspace serves as the primary carrier for improving reasoning performance. Remarkably, OPD consistently outperforms RL across all evaluated rank levels, and this advantage persists across different model scales and rank thresholds. This suggests not only that OPD allocates its limited update budget more efficiently by concentrating on higher-quality directional subspaces, but also that the principal directions identified by OPD inherently encode more effective update signals than those of RL, even under the same norm budget.

\vskip -0.2in
\paragraph{Bottom-$k\%$ Subspace: Marginal Utility of Tail Directions.}

\begin{figure}[t]
    \includegraphics[width=1\textwidth]{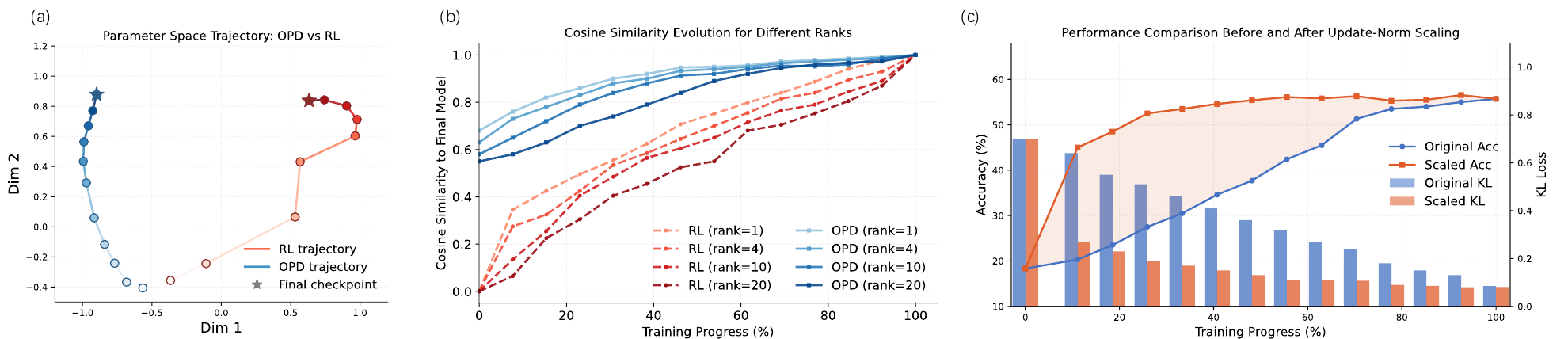}
    \caption{Subspace evolution and weight scaling analysis during training. (a) t-SNE visualization of Top-1 subspace evolution for RL and OPD trajectories. (b) Cosine similarity between the Top-$k$ subspaces of intermediate and final checkpoints. (c) Changes in Accuracy and KL after scaling intermediate OPD checkpoints' $\Delta W_{\text{OPD}}$ to match the final checkpoint's norm.}
    \label{fig4}
\end{figure}

To further investigate, we compare the impact of tail directions on performance, where tail directions are defined as the subspace constructed using the last $k\%$ singular components, denoted as $\Delta W_{\text{Bottom-}k\%}$. Unlike the Top-$k\%$ subspace analysis, we do not apply norm scaling to equalize the update budgets, so as to observe their performance contributions under the original training state. As shown in Figure~\ref{fig4} (b), in contrast to the principal subspace, tail subspaces provide only limited performance recovery for both RL and OPD. On the Qwen2.5-1.5B-DeepSeek model, retaining only 10\% of the principal subspace increases reasoning accuracy from 23.33\% to 40.3\%, whereas preserving 50\% of the tail subspace achieves only around 30\%, despite using a much larger fraction of the rank budget. This contrast suggests that tail directions have substantially lower marginal utility for reasoning than principal directions.

Interestingly, RL exhibits a slight advantage over OPD in tail directions. However, this marginal benefit comes with a large norm cost: the norm of RL's tail subspace ($\Delta W_{\text{Bottom-}50\%}$) ranges from approximately 1.6 to 2.5 times that of OPD, while the corresponding performance gain remains limited. In other words, RL allocates a substantial portion of its update magnitude to tail directions, but the marginal return of this allocation is relatively low.

These observations help explain the compactness advantage of OPD discussed in Section~\ref{section2}. Compared with OPD, RL distributes more update energy into tail directions whose contribution to reasoning performance is limited, which is consistent with its larger overall update norm for comparable performance. In contrast, OPD allocates a larger fraction of its update energy to the principal subspace, thereby achieving stronger per-norm performance gains with more compact updates.

The preceding analysis shows that OPD updates exhibit substantially stronger low-rank concentration from a spatial-geometric perspective. Together with the controlled Top-$k\%$ and Bottom-$k\%$ subspace experiments, this suggests that such concentration is a key factor behind OPD's higher per-norm efficiency, rather than merely a by-product of smaller update norms. We next move from static spectral structure to temporal evolution, examining whether OPD's efficiency arises from early identification of high-quality directions or from continuous path correction during training.

\subsection{Directional Stabilization and Magnitude Development}
\label{Directional Stabilization and Magnitude Development}
\paragraph{Subspace Evolution Trajectory Analysis.}
To qualitatively compare the evolution of update directions during training, we visualize the Top-1 subspace using t-SNE, as shown in Figure~\ref{fig4} (a). The RL trajectory exhibits larger variations across checkpoints, whereas the OPD trajectory appears more compact and smoother in the projected space. This visualization suggests a potential difference in directional stability between RL and OPD, which we next examine quantitatively through subspace alignment analysis.

Specifically, we pair each Top-$k$ subspace ($k=1,\ldots,20$) from each training step with its corresponding subspace in the final checkpoint, compute the cosine similarity, and then average over $k$. The results are shown in Figure~\ref{fig4} (b). OPD consistently exhibits stronger alignment with its final subspaces than RL across all evaluated ranks, with smaller fluctuations throughout training. This difference is particularly pronounced in the early stage of training (0\%--30\%), indicating that OPD stabilizes its dominant update directions earlier than RL, and that this stability extends beyond the Rank-1 direction to multiple dominant subspaces.

\paragraph{Magnitude Scaling and Performance Recovery.}
The preceding subspace-alignment analysis shows that the dominant OPD update subspaces are already strongly aligned with their final counterparts at an early stage of training. Based on this observation, we further investigate the source of the remaining performance gap in early checkpoints: whether this gap arises from insufficiently formed effective update directions, or from underdeveloped update magnitudes along these directions.

To examine this hypothesis, we perform a module-wise norm-scaling intervention on intermediate OPD checkpoints. For each intermediate checkpoint, we preserve the update direction within each module, while rescaling its Frobenius norm to match that of the corresponding module in the final checkpoint. We then apply the rescaled update to the base model and evaluate the resulting model, as shown in Figure~\ref{fig4} (c). This intervention allows us to assess how much performance can be recovered when early update directions are given the same module-wise norm budget as the final checkpoint.

The results show that norm scaling markedly improves the performance of early checkpoints. In particular, a checkpoint at only 10\% training progress recovers approximately 80\% of the final model's performance after scaling. We also observe a reduction in the KL divergence between the rescaled checkpoints and the teacher model, indicating that the scaled updates move the student output distribution closer to the teacher distribution. These results suggest that early OPD checkpoints already possess task-relevant update directions, while the limited update magnitudes become a bottleneck that constrains further performance improvement.

Overall, these experiments separate two aspects of the OPD update trajectory, namely the formation of dominant directions and the growth of update magnitudes, thereby complementing the subspace alignment analysis. Experimental evidence shows that OPD establishes stable update directions early in training, with subsequent training primarily accumulating magnitude along these directions rather than making large-scale adjustments to the directions themselves. We further analyze the geometric and theoretical manifestations of Property 2 in Appendix~\ref{Cosine Similarity Analysis of Subspace}-\ref{A Linearized View of OPD Dynamics}.

\paragraph{Summary.} This section reveals the core geometric characteristics of OPD's parameter updates. OPD's updates exhibit stronger low-rank concentration and stabilize their dominant subspaces early, with subsequent training mainly progressing along these subspaces. We term this \textbf{Property~2: Early Low-Rank Lock-in}, which structurally explains \textbf{Property~1: Functional Redundancy Avoidance}. By locking into efficient low-rank directions early, OPD reduces reliance on redundant exploration and correction, avoids overlearning redundant information, and exhibits stronger foresight at the modular level.

\begin{figure}[t]
    \includegraphics[width=1\textwidth]{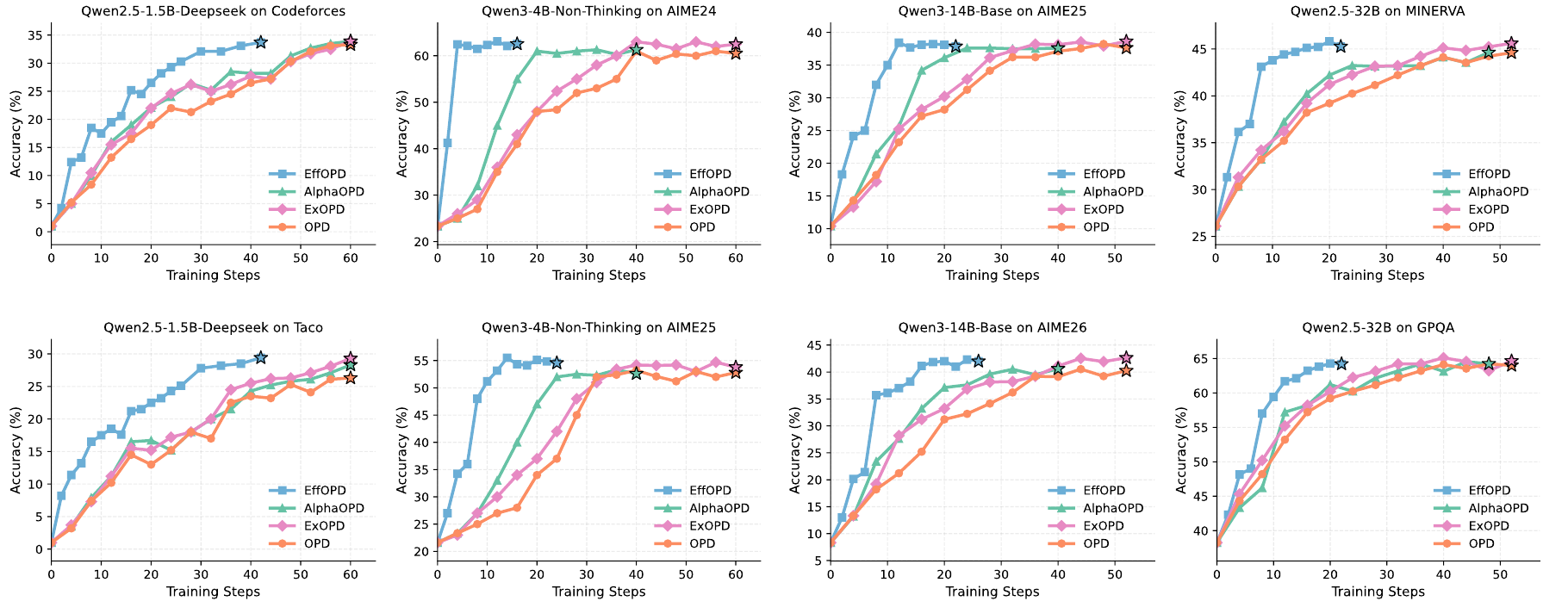}
    \caption{Performance comparison of different distillation methods on code and math datasets.}
    \label{fig5}
\end{figure}

\section{Accelerating OPD via Directional Extrapolation}
The preceding analysis suggests that OPD establishes highly stable and final-aligned update directions early in training. After this early directional lock-in, later optimization mainly amplifies the update magnitude along the same trajectory, rather than exploring new directions. Motivated by this observation, we propose \textbf{EffOPD}, a plug-and-play acceleration framework that exploits early directional extrapolation to accelerate OPD. We next detail the acceleration procedure and report the corresponding empirical results.

\subsection{Method}

Let $W_t$ denote the model parameters after the $t$-th OPD update. EffOPD triggers an extrapolation search at exponentially spaced checkpoints, i.e., when $t=2^n$ with $n$ starting from 0, so the first extrapolation is performed at $t=1$. For the first checkpoint, we use the displacement from the initial parameters to $W_1$ as the local update direction. For subsequent checkpoints with $n \ge 1$, EffOPD estimates the local update direction using the parameter displacement between the current exponential checkpoint and the previous one: 
\begin{equation}
\Delta_n = W_{2^n} - W_{2^{n-1}}.
\end{equation}

This displacement captures the accumulated parameter evolution between two adjacent exponential checkpoints. Since OPD update directions remain relatively stable during training, $\Delta_n$ serves as a local approximation of subsequent update directions.

EffOPD then generates five candidate parameters from $W_{2^n}$ along $\Delta_n$ with increasing extrapolation magnitudes. For $k=1,2,\cdots,5$, the $k$-th candidate is defined as:
\begin{equation}
\widetilde{W}_{n,k}
=
W_{2^n}
+
{2k}\Delta_n,
\end{equation}
where the coefficient ${2k}$ controls the extrapolation scale. To determine whether the extrapolated parameters remain effective, EffOPD randomly samples 50 examples from the training set to form a lightweight validation set $\mathcal{D}_v$, which is far smaller than the number of sentences generated per step in vanilla OPD. sLet $\mathcal{V}_{\mathcal{D}_v}(\cdot)$ denote the validation function. EffOPD initializes the accepted parameters as $W^{\mathrm{acc}}=W_{2^n}$ and its score as $v^{\mathrm{acc}}=\mathcal{V}_{\mathcal{D}_v}(W_{2^n})$. Then EffOPD evaluates $\widetilde{W}_{n,k}$ sequentially. If $\mathcal{V}_{\mathcal{D}_v}(\widetilde{W}_{n,k}) \ge v^{\mathrm{acc}}$, the candidate is accepted, and we update:
\begin{equation}
W^{\mathrm{acc}} \leftarrow \widetilde{W}_{n,k}, \quad v^{\mathrm{acc}} \leftarrow \mathcal{V}_{\mathcal{D}_v}(\widetilde{W}_{n,k}).
\end{equation}
If the current candidate fails to improve validation performance, the search terminates immediately. Thus, the final accepted parameters $W_{2^n}^{\mathrm{EffOPD}}$ at checkpoint $2^n$ is:
\begin{equation}
W_{2^n}^{\mathrm{EffOPD}} = W^{\mathrm{acc}}.
\end{equation}
In particular, if the candidate with $k=1$ already fails, EffOPD degenerates to vanilla OPD. This progressive extrapolation and immediate validation mechanism enables EffOPD to exploit the early directional stability of OPD while avoiding performance degradation caused by excessive extrapolation.

\begin{figure}[t]
    \includegraphics[width=1\textwidth]{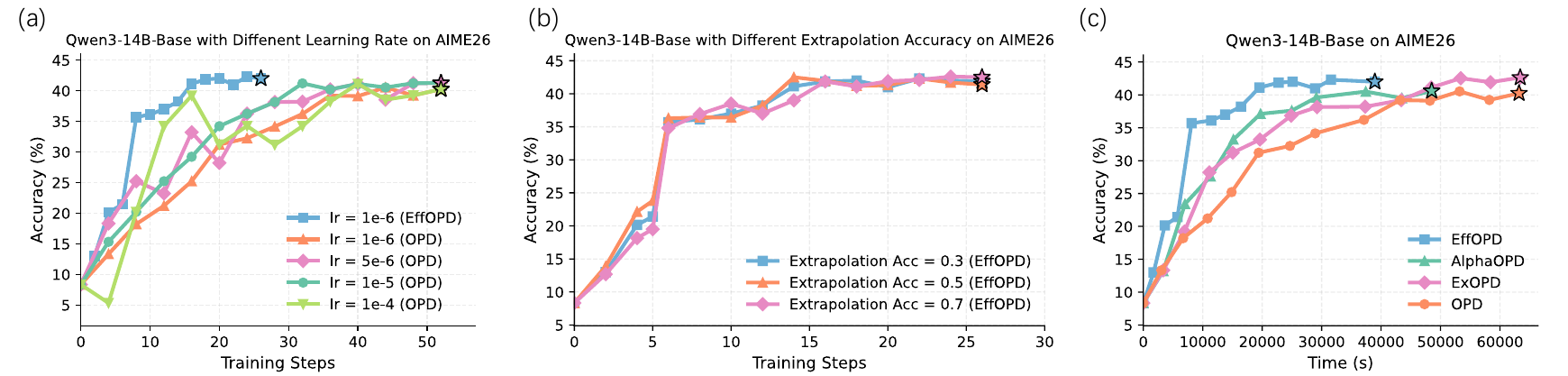}
    \caption{Ablation studies. (a) Effect of different learning rates. (b) Impact of $\mathcal{D}_v$ difficulty on EffOPD. ``Extrapolation Acc'' denotes the accuracy of the model before training on the sampled $\mathcal{D}_v$. (c) Relationship between training time and performance.}
    \label{fig6}
\end{figure}

\subsection{Main Results}
To evaluate EffOPD, we conduct experiments on code generation and mathematical reasoning. We use Eurus-RL-Code~\citep{cui2025processreinforcementimplicitrewards} and DeepMath-103K~\citep{Yang2026LearningBT} for training, and evaluate models at four scales: 1.5B, 4B, 14B, and 32B. For each scale, the RL-finetuned model serves as the teacher. We report results on seven benchmarks: Codeforces, Taco~\citep{liu2023codegeneratedchatgptreally}, AIME24, AIME25, AIME26, MINERVA, and GPQA \citep{ye2025limoreasoning}. We compare EffOPD with Vanilla OPD, AlphaOPD~\citep{cai2025predictability}, and ExOPD~\citep{Yang2026LearningBT}.

As shown in Figure~\ref{fig5}, EffOPD consistently improves training efficiency across all model scales and datasets. On mathematical reasoning tasks, it typically begins to converge within about 10 training steps, compared with 30--40 steps for vanilla OPD, yielding more than a $3\times$ speedup. EffOPD also reaches a higher performance upper bound, possibly because prolonged vanilla OPD training may cause over-optimization and semantic drift. Unlike AlphaOPD and ExOPD, which use fixed extrapolation strategies, EffOPD adaptively selects the extrapolation magnitude via validation feedback, leading to more stable acceleration. Its early-stage advantage is especially evident on Qwen3-4B-Non-Thinking, where EffOPD attains strong reasoning performance by the 4th step, further supporting that OPD forms high-quality, well-aligned update directions early in training.

\textbf{Ablation Studies.}
We conduct ablation studies to identify the key factors behind EffOPD's effectiveness. As shown in Figure~\ref{fig6} (a), the learning rate strongly affects the stability of vanilla OPD: larger learning rates accelerate early convergence but also cause noticeable oscillations and performance instability. In contrast, EffOPD uses lightweight validation during extrapolation to adaptively filter out overly aggressive steps, thereby improving training stability. Figure~\ref{fig6} (b) shows that the difficulty of the lightweight validation set $\mathcal{D}_v$ is not critical. Validation sets of different difficulty levels provide consistent directional signals, suggesting that validation mainly serves to check whether the current update direction remains effective rather than to provide precise supervision. Figure~\ref{fig6} (c) compares actual training time. Despite the additional validation overhead, EffOPD achieves better performance under the same time budget and converges faster than vanilla OPD, indicating that the gain from exploiting early-stage update directions outweighs the validation cost. Overall, these results support the proposed foresight mechanism: once OPD establishes effective directions early in training, EffOPD can safely extrapolate along them to achieve stable and efficient acceleration.

\section{Conclusion}
In this work, we identify two properties that reveal the underlying ``foresight'' of OPD: \textbf{Functional Redundancy Avoidance} at the modular level and \textbf{Early Low-Rank Lock-in} at the update-direction level. Through parameter-level analyses across model scales, RL algorithms, and task domains, we show that OPD achieves RL-comparable reasoning gains with more compact and structured updates, as it concentrates optimization on high-utility modules and directions from the early stage of training. Building on this insight, we propose \textbf{EffOPD}, a plug-and-play acceleration method that leverages early directional stability to achieve up to \(3\times\) training speedup while maintaining the final performance. Overall, our findings suggest that OPD's efficiency is fundamentally tied to early directional stabilization and compact parameter allocation, offering a new perspective for understanding and accelerating post-training in large language models.


\bibliography{nips2026}
\bibliographystyle{nips2026}

\appendix

\newpage

\section{Impact Statement}
This paper presents work whose goal is to advance the understanding and efficiency of post-training for large language models, particularly on-policy distillation. We believe that our work conforms with the NeurIPS Code of Ethics. The proposed analysis and EffOPD method may help reduce the computational cost of post-training and make efficient model improvement more accessible. However, more efficient post-training techniques could also be misused to enhance or adapt models for harmful applications. We encourage responsible use of these methods, together with appropriate safety evaluation and deployment safeguards.

\section{Related Work}
\label{Related Work}

\textbf{On-policy Distillation (OPD).}
In this paradigm, the student generates its own samples and receives dense supervisory signals from the teacher \citep{agarwal2024policy}. Qwen3 \citep{yang2025qwen3technicalreport} demonstrates that it achieves substantially higher training efficiency than RLVR. Meanwhile, MiMo-V2-Flash \citep{xiao2026mimo} and Deepseek-V4 \citep{deepseek2026v4} integrate multiple teacher skills into a small model via multi-task on-policy distillation. \cite{song2026survey} present the first systematic survey of OPD for large language models, proposing a unified $f$-divergence framework grounded in on-policy samples. \cite{fu2026revisiting} prove that token-level OPD is biased relative to the sequence-level reverse-KL objective but has a tighter variance bound of $O(T^2)$ versus $O(T^4)$. \cite{yang2026learning} establish a theoretical equivalence between token-level distillation and RLVR. \cite{li2026rethinking} systematically investigate the training dynamics of OPD and identify two necessary conditions for success: (i) the student and teacher must share compatible thinking patterns, and (ii) the teacher must offer genuinely novel capabilities beyond what the student has encountered during training.

\textbf{Emergent Behaviors of On-Policy Training.}
\cite{yue2025doesreinforcementlearningreally} investigated the differences in sampling between base models and RL-fine-tuned models, showing that RL improves sampling efficiency for pass@1 but does not directly enhance reasoning ability.  
\cite{cui2025entropymechanismreinforcementlearning} identified the phenomenon of ``entropy collapse'' in reinforcement learning, where rapid early convergence causes the model to become overly confident, prematurely degrading its exploratory capacity. Through systematic experiments across models of varying scales, \cite{tan2026scalingbehaviorsllmreinforcement} reveal a power-law relationship between test loss, computational budget, and data volume during RL post-training of LLMs, demonstrating that larger models consistently exhibit superior learning efficiency. \cite{cai2025predictability} investigate RL from the perspective of parameter dynamics. They uncover two fundamental properties of RL-induced updates: Rank-1 dominance and Rank-1 linear dynamics. Based on these insights, their AlphaRL framework achieves $3 \times$ training acceleration. Building on this, \cite{chen2026lowrankoptimizationtrajectoriesmodeling} train a predictor that directly forecasts the evolution direction of subsequent optimization subspaces using the early Rank‑1 subspace. Different from previous studies focusing on RL's low-rank trajectories, this work finds that OPD's efficiency advantage over RL stems from the unique synergy between modular redundancy suppression and early directional stabilization.

\section{Limitations and Future Work}
\label{Limitation}

Despite our identification of two properties of OPD, this study has several limitations. First, although these properties are validated from multiple perspectives, their applicability to more complex settings, such as multi-turn agent tasks and multimodal reasoning, remains to be further examined. These settings may introduce stronger distributional shifts and more complex teacher-student residual structures. Second, our theoretical analysis in Appendix is inherently local, characterizing OPD dynamics only in a neighborhood of the base model and therefore not fully capturing the global non-convex behavior of large-scale post-training.

These limitations point to several directions for future work. A more complete theory should account for the coupling between the distillation objective, the evolving on-policy distribution, and the spectral evolution of parameter updates. In addition, the early directional lock-in observed in OPD may serve as a useful diagnostic signal for monitoring post-training dynamics. Metrics such as directional alignment, spectral concentration, and update compactness could help assess training progress and stability, thereby supporting more adaptive and efficient on-policy distillation methods for large language models.

\clearpage
\section{Preliminaries and Experimental Setup}
\label{Preliminaries and Experimental Setup}

\subsection{Preliminaries}
In our experiments, we focus on the two training paradigms: Reinforcement Learning \citep{Zhang2025ASO} and On-Policy Distillation \citep{Kim2026WhyDS}. Let $\pi_{\theta}$ denote the policy model to be optimized.

\paragraph{Reinforcement Learning (RL).}
The RL objective can be formulated as:
\begin{equation}
J_{\text{RL}}(\theta) = \max_{\theta} \; \mathbb{E}_{x \sim \mathcal{D}, \; y \sim \pi_{\theta}(\cdot \mid x)} \left[ r(x, y) - \beta D_{\text{KL}}\bigl(\pi_{\theta} \parallel \pi_{\text{ref}}\bigr) \right],
\label{eq:rl_objective}
\end{equation}
where the trajectory $y = (y_1, \ldots, y_T)$ is sampled from the current policy $\pi_{\theta}$, ensuring on-policy training. The function $r(x, y)$ measures the quality of response $y$ to query $x$. In the Reinforcement Learning from Verifiable Rewards setting (RLVR) \citep{Venkatkrishna2026AletheiaWM}, $r(x, y)$ is a deterministic verifiable reward (e.g., answer correctness or unit test passing), requiring no learned reward model. The term $D_{\text{KL}}(\pi_{\theta} \parallel \pi_{\text{ref}})$ is a KL constraint that prevents the policy from deviating too far from a reference model $\pi_{\text{ref}}$, with $\beta$ controlling the constraint strength.

To optimize Eq.~\eqref{eq:rl_objective}, policy gradient methods are commonly used, yielding the following gradient estimate:
\begin{equation}
\nabla_{\theta} J_{\text{RL}}(\theta) = \mathbb{E}_{x \sim \mathcal{D}, \; y \sim \pi_{\theta}(\cdot \mid x)} \left[ \sum_{t=1}^{T} A_t \nabla_{\theta} \log \pi_{\theta}(y_t \mid x, y_{<t}) \right],
\label{eq:rl_gradient}
\end{equation}
where $A_t$ is the advantage of token $y_t$ relative to a baseline. In practice, the reward signal in RLVR is often sparse, as the policy only receives a reward upon completion of the full response.

\paragraph{On-Policy Distillation (OPD).}
OPD inherits the on-policy nature of policy training while leveraging dense supervisory signals from a teacher model, making it an efficient post-training paradigm \citep{Yang2026LearningBT}. The core idea is to let the student model $\pi_{\theta}$ generate its own trajectories $y$, and then minimize the reverse KL divergence between the student and a fixed teacher model $\pi^*$ on these student-generated trajectories:
\begin{equation}
J_{\text{OPD}}(\theta) = \min_{\theta} \; \mathbb{E}_{x \sim \mathcal{D}, \; y \sim \pi_{\theta}(\cdot \mid x)} \left[ D_{\text{KL}}\bigl(\pi_{\theta}(y \mid x) \parallel \pi^*(y \mid x)\bigr) \right].
\label{eq:opd_objective}
\end{equation}
Note that the trajectories $y$ in Eq.~\eqref{eq:opd_objective} are sampled from the student policy $\pi_{\theta}$ itself, preserving the on-policy property. The corresponding gradient is:
\begin{equation}
\nabla_{\theta} J_{\text{OPD}}(\theta) = \mathbb{E}_{x \sim \mathcal{D}, \; y \sim \pi_{\theta}(\cdot \mid x)} \left[ \sum_{t=1}^{T} \sum_{t'=t}^{T} \Bigl( \log \pi_{\theta}(y_{t'} \mid x, y_{<t'}) - \log \pi^*(y_{t'} \mid x, y_{<t'}) \Bigr) \nabla_{\theta} \log \pi_{\theta}(y_t \mid x, y_{<t}) \right].
\label{eq:opd_gradient_full}
\end{equation}

In practice, following prior work, a common approximation sets the discount factor to zero, focusing on immediate token-level optimization:
\begin{equation}
\nabla_{\theta} J_{\text{OPD}}(\theta) \approx \mathbb{E}_{x \sim \mathcal{D}, \; y \sim \pi_{\theta}(\cdot \mid x)} \left[ \sum_{t=1}^{T} \Bigl( \log \pi_{\theta}(y_t \mid x, y_{<t}) - \log \pi^*(y_t \mid x, y_{<t}) \Bigr) \nabla_{\theta} \log \pi_{\theta}(y_t \mid x, y_{<t}) \right].
\label{eq:opd_gradient_approx}
\end{equation}

This approximation provides a dense learning signal at every token position, enabling OPD to achieve significantly higher training efficiency compared to RLVR with its sparse reward signal.

\newpage
\subsection{Experimental Setup}
\label{Experimental Setup}

\begin{table}[h]
\centering
\caption{Summary of models considered in this study.}
\setlength{\tabcolsep}{4pt}
\label{mode_config}
\begin{tabular}{ccccc}
\toprule
\textbf{Base Model} & \textbf{RL Model} & \textbf{Algorithm} & \textbf{Open-Source} \\
\midrule
Qwen2.5-1.5B-Deepseek & JustRL \citep{he2025justrlscaling15bllm} & GRPO & Yes \\
Qwen2.5-1.5B-Deepseek & BroRL \citep{Hu2025BroRLSR} & PPO & Yes \\
Qwen2.5-1.5B-Deepseek & ProRL \citep{Liu2025ProRLPR} & DAPO & Yes \\
Qwen3-4B-Non-Thinking & Qwen-4B-Non-Thinking-GRPO & GRPO & Yes \\
Qwen2.5-7B & Open-Reasoner-Zero \citep{hu2025openreasonerzeroopensourceapproach} & PPO & Yes \\
Qwen3-8B-Base & Qwen3-8B-PPO \citep{cai2025predictability} & PPO & Yes \\
Qwen3-8B-Base & Qwen3-8B-DAPO \citep{cai2025predictability} & DAPO & Yes \\
Qwen3-14B-Base & Qwen3-14B-Base-DAPO & DAPO & No \\
Qwen3-14B & Qwen3-14B-GRPO & GRPO & No \\
Qwen2.5-32B & DAPO-Qwen-32B \citep{yu2025dapoopensourcellmreinforcement} & DAPO & Yes \\
\bottomrule
\end{tabular}
\end{table}

To ensure the generality of our findings, we conduct experiments across multiple model scales, ranging from 1.5B to 32B parameters. Our experimental models include publicly available pre-trained checkpoints (e.g., Qwen2.5-7B, Qwen3-4B, etc.), as well as models locally trained using the Verl framework. For RL methods, we consider three representative algorithms—PPO, GRPO, and DAPO—and apply them to models of varying scales. For all OPD student models reported in Table \ref{mode_config}, the capability-aligned teacher is consistently the RL-tuned version of its own base model (i.e., the RL model listed in the same table); for Qwen3-8B-Base, we also use Qwen3-14B-Base-DAPO as the teacher to ensure the generality of our conclusions.

For models trained with reinforcement learning locally, we adapt our training codebase using \texttt{Verl}~\citep{Sheng_2025} and follow the corresponding training setups. All methods share the same core configuration: the maximum prompt length is 2{,}048 tokens and the maximum response length is 20{,}480 tokens, yielding a total budget of 22{,}528 tokens. During training, each backward pass uses a mini-batch of 32 samples, and gradients are accumulated for 16 iterations before a single optimization step is performed, resulting in an effective batch size of 512 under \texttt{Float16} precision. Each prompt generates $n = 16$ outputs during rollout. The learning rate is set to $1 \times 10^{-6}$ with warmup, and gradient clipping of 1.0 is applied. We monitor the average reward per training batch and terminate training once the reward fails to improve for five consecutive steps.

In addition to the unified configuration described above, each method adopts specific hyperparameter settings in our experiments. For \textbf{GRPO}, we set both the high and low clipping ratios to 0.2 and apply a KL loss with coefficient $0.001$, following \cite{deepseekai2025deepseekr1incentivizingreasoningcapability}. For \textbf{DAPO}, we employ techniques such as clip-higher, dynamic sampling, token-level policy gradient loss, and overlong reward shaping, and apply the recommended hyperparameters from \cite{yu2025dapoopensourcellmreinforcement}: the clipping ratios are set to $\epsilon_{\text{low}} = 0.2$ and $\epsilon_{\text{high}} = 0.28$, KL divergence terms are removed entirely. We perform RLVR training on Qwen3-14B-Base models using the DeepMath-103K~\citep{he2025deepmath103klargescalechallengingdecontaminated} and MATH-12K \citep{lightman2023lets} for training. For the Qwen3-14B models, we conduct rollout and training in their non-thinking mode and we employ the built-in chat template, specified as follows:

\begin{verbatim}
User: 
{question} 
Please reason step by step, and put your final answer within \boxed{}.
<think>
</think>
Assistant: {CoT}
\end{verbatim}

For OPD, we follow the setting of \citet{Yang2026LearningBT}. The maximum prompt length is 2{,}048 tokens and the maximum response length is 16{,}384 tokens, yielding a total budget of 18{,}432 tokens. The prompt batch size is 1{,}024, and each prompt generates $n = 1$ outputs during rollout. The learning rate is set to $1 \times 10^{-6}$, without warmup, and a total of 3 training epochs. The next page shows the OPD training command using the \texttt{verl} framework. All of our training runs are conducted on $8\times$ or $32\times$ H20 96GB GPUs.

\begin{tcolorbox}[
    colback=gray!5,          
    colframe=gray!70,        
    arc=3pt,                 
    left=2pt,
    right=2pt,
    top=2pt,
    bottom=2pt,
    boxrule=0.5pt,           
    breakable,               
    fontupper=\small,
    listing only,            
    listing options={language=Python}, 
    title=OPD Training Command, 
    title style={color=black},          
    label=py1                
]
\begin{verbatim}

python3 -m verl.trainer.main_ppo \
    algorithm.adv_estimator=grpo \
    algorithm.rollout_correction.rollout_is=token \
    algorithm.rollout_correction.rollout_is_threshold=5.0 \
    algorithm.rollout_correction.rollout_rs=null \
    algorithm.rollout_correction.bypass_mode=false \
    actor_rollout_ref.rollout.calculate_log_probs=true \
    data.train_files=/path/to/train.parquet \
    data.val_files=/path/to/val.parquet \
    data.train_batch_size=1024 \
    data.max_prompt_length=2048 \
    data.max_response_length=16384 \
    data.filter_overlong_prompts=True \
    data.truncation='error' \
    data.shuffle=True \
    data.seed=42 \
    data.return_raw_chat=True \
    +data.apply_chat_template_kwargs.enable_thinking=False \
    actor_rollout_ref.model.path=$MODEL_PATH \
    +actor_rollout_ref.ref.model.path=$TEACHER_MODEL_PATH \
    actor_rollout_ref.actor.optim.lr=1e-6 \
    actor_rollout_ref.actor.optim.lr_warmup_steps_ratio=0.0 \
    actor_rollout_ref.model.use_remove_padding=True \
    actor_rollout_ref.actor.policy_loss.only_reverse_kl_advantages=True \
    actor_rollout_ref.actor.ppo_mini_batch_size=1024 \
    actor_rollout_ref.actor.ppo_micro_batch_size_per_gpu=1 \
    actor_rollout_ref.actor.use_kl_loss=True \
    actor_rollout_ref.actor.kl_loss_coef=0 \
    actor_rollout_ref.actor.kl_loss_type=low_var_kl \
    actor_rollout_ref.actor.entropy_coeff=0 \
    actor_rollout_ref.actor.ppo_max_token_len_per_gpu=22000 \
    actor_rollout_ref.model.enable_gradient_checkpointing=True \
    actor_rollout_ref.actor.fsdp_config.param_offload=False \
    actor_rollout_ref.actor.fsdp_config.optimizer_offload=False \
    actor_rollout_ref.rollout.log_prob_micro_batch_size_per_gpu=1 \
    actor_rollout_ref.rollout.tensor_model_parallel_size=1 \
    actor_rollout_ref.rollout.name=vllm \
    actor_rollout_ref.rollout.gpu_memory_utilization=0.6 \
    actor_rollout_ref.rollout.n=1 \
    actor_rollout_ref.rollout.max_num_batched_tokens=22000 \
    actor_rollout_ref.rollout.temperature=1.0 \
    actor_rollout_ref.rollout.top_p=1.0 \
    actor_rollout_ref.rollout.val_kwargs.do_sample=True \
    actor_rollout_ref.rollout.val_kwargs.temperature=1.0 \
    actor_rollout_ref.rollout.val_kwargs.top_p=1.0 \
    actor_rollout_ref.rollout.val_kwargs.n=32 \
    actor_rollout_ref.ref.log_prob_micro_batch_size_per_gpu=1 \
    actor_rollout_ref.ref.fsdp_config.param_offload=True \
    algorithm.use_kl_in_reward=False \
    reward_model.reward_manager=naive \
    trainer.critic_warmup=0 \
    trainer.val_before_train=True \
    trainer.logger='["console","wandb"]' \
    trainer.log_val_generations=10 \
    trainer.project_name='on-policy-distillation' \
    trainer.experiment_name='on-policy-distillation' \
    trainer.n_gpus_per_node=8 \
    trainer.nnodes=4 \
    trainer.save_freq=2 \
    trainer.default_local_dir=/path/to/save/dir \
    trainer.test_freq=2 \
    trainer.total_epochs=3 $@
\end{verbatim}
\end{tcolorbox}

\newpage

\section{Property 1 Additional Experiment}
\label{Property 1 Additional Experiment}

\subsection{Additional Experiment}
\label{Additional Experiment}
This section provides additional empirical evidence to further validate \textbf{Property 1 (Functional Redundancy Avoidance)} introduced in Section~\ref{section2}. 

\begin{figure}[t]
    \includegraphics[width=0.98\textwidth]{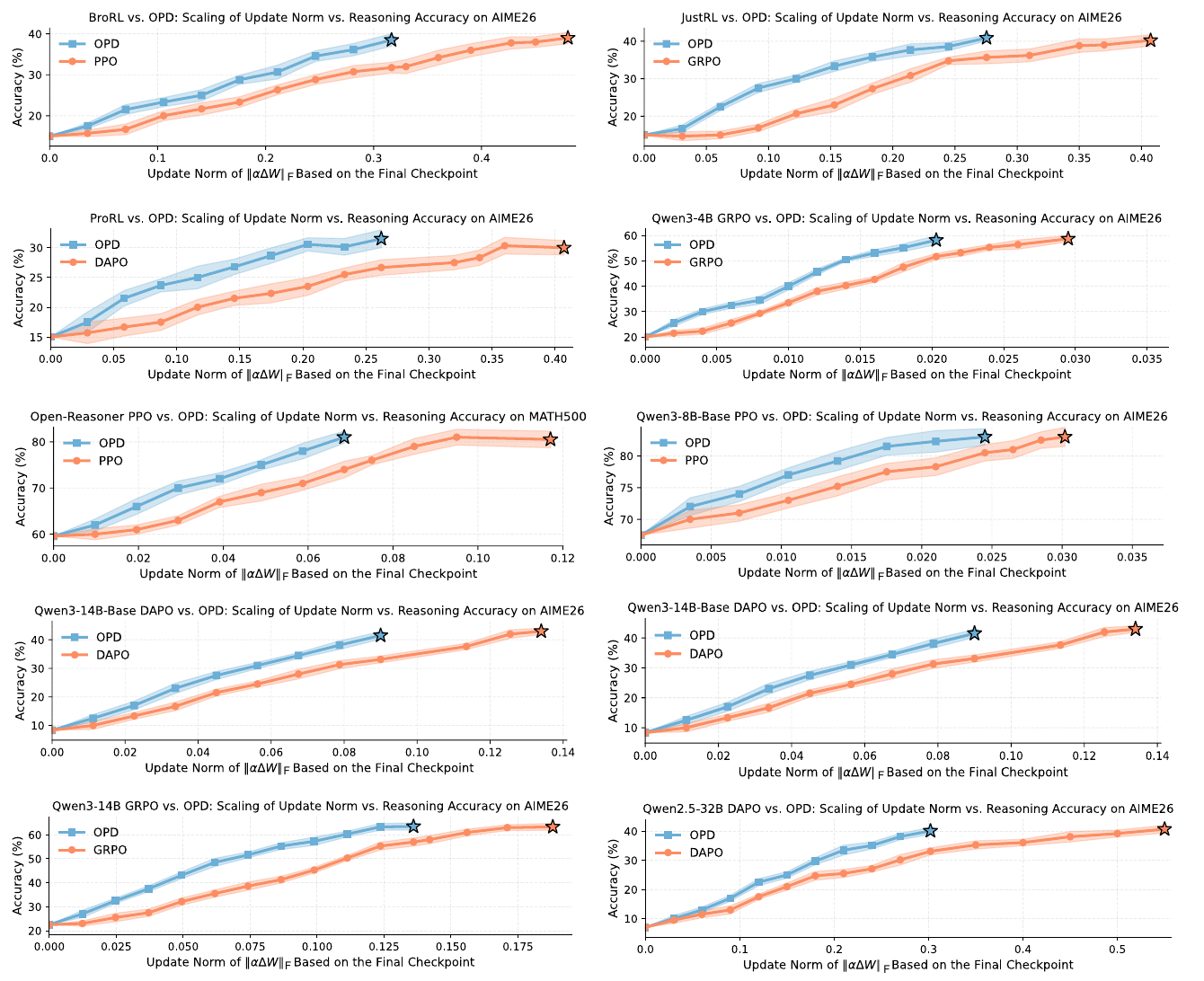}
    \caption{Comparison of parameter update efficiency between RL and OPD. Scaling analysis of the final checkpoints demonstrates that OPD achieves substantially higher reasoning gains than RL under an identical update norm budget.}
    \label{appendixfig1}
\end{figure}

We begin by examining the scaling behavior across model sizes. Figure~\ref{appendixfig1} presents the scaling results on final checkpoints for models ranging from 1.5B to 32B parameters. Across all scales, we observe a consistent pattern: OPD achieves reasoning performance comparable to that of RL while requiring substantially smaller parameter update norms. This result suggests that the functional efficiency of OPD is not a scale-specific artifact, but rather an intrinsic property that generalizes across model sizes. We attribute this behavior to OPD's ability to systematically suppress functionally redundant updates, thereby concentrating the update budget on more effective directions.

\begin{figure}[t]
    \includegraphics[width=0.99\textwidth]{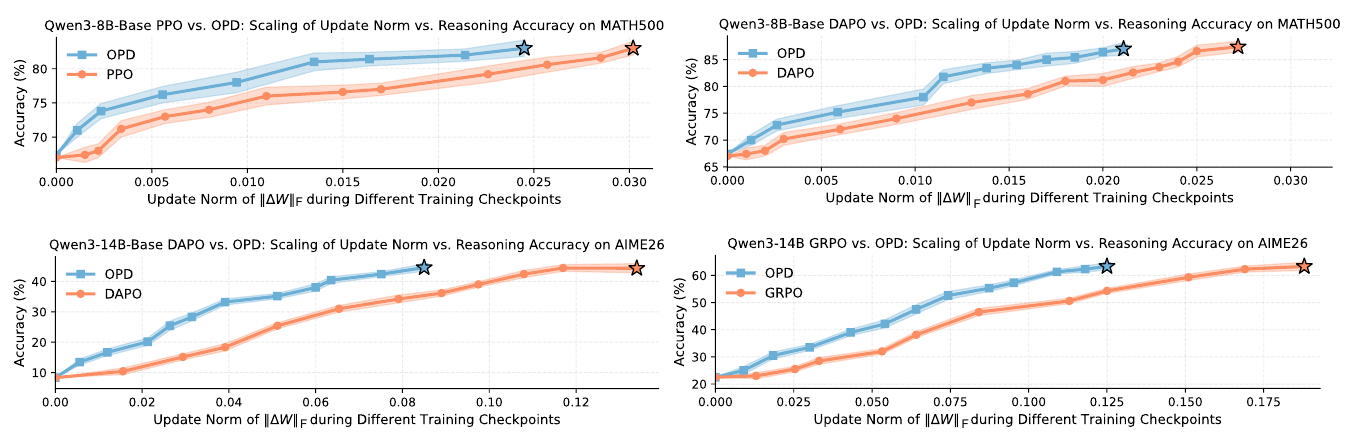}
    \caption{Comparison of parameter update efficiency between RL and OPD. Analysis of intermediate checkpoints throughout training demonstrates that OPD achieves the same reasoning accuracy as RL with substantially smaller parameter update norms.}
    \label{appendixfig2}
\end{figure}

We next investigate whether this advantage persists across different reinforcement learning algorithms. Figure~\ref{appendixfig2} extends the analysis to a broader set of RL methods. Across all examined algorithms, OPD consistently demonstrates superior parameter update efficiency throughout the training trajectory. This advantage holds regardless of the specific learning dynamics or convergence behavior of the teacher RL method, indicating that the efficiency gain arises from the structural properties of OPD updates rather than the choice of the underlying RL algorithm. Taken together, these results provide consistent cross-scale and cross-algorithm evidence that OPD achieves comparable or even superior reasoning performance with significantly improved parameter efficiency.

While the main text shows that embedding layer updates contribute negligibly to reasoning performance, it does not explicitly analyze their distributional shift relative to the base model. To address this, we sample reasoning sequences generated by the base model and extract their token embeddings. We then visualize the embedding shifts using PCA \citep{Eckart1936TheAO} and t-SNE \citep{shi2021visualizingdatausinggtsne}, and quantify the distributional differences via cosine similarity between token representations. As shown in Figure~\ref{appendix4} and Table~\ref{embed_similarity}, OPD consistently exhibits smaller embedding shifts than RL across all model scales, and maintains higher similarity to the base representations. These findings indicate that, despite their limited functional contribution, embedding layers in OPD still undergo more constrained and compact updates, effectively avoiding the unnecessary drift commonly observed in RL. This suggests that OPD enforces compact updates not only in critical modules but also in functionally peripheral regions.

\begin{figure}[t]
    \includegraphics[width=1\textwidth]{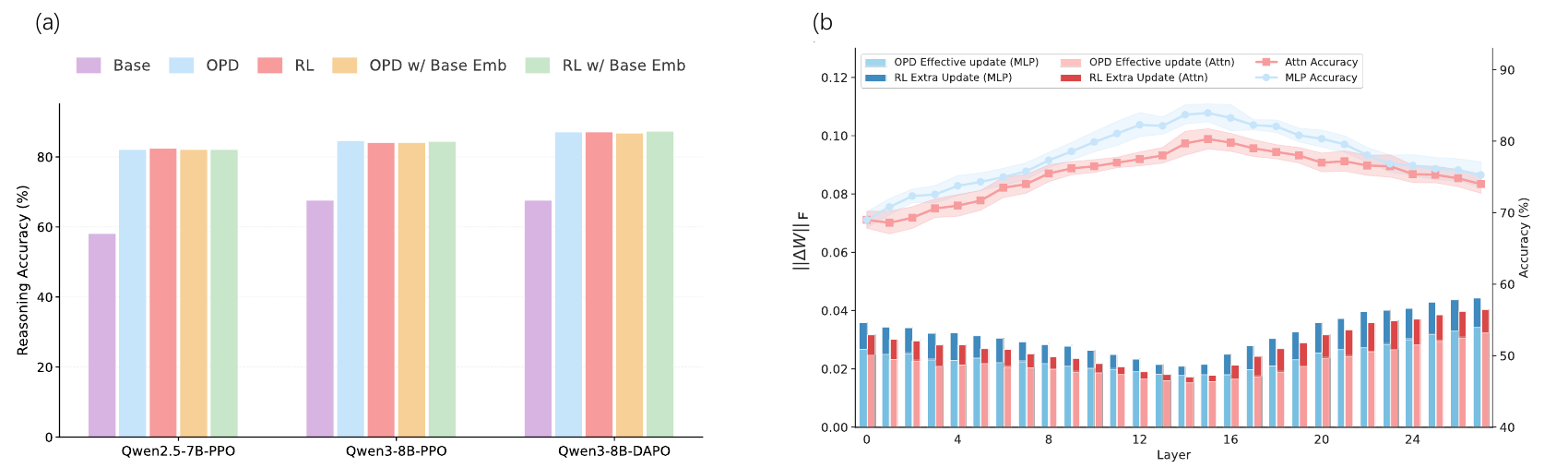}
    \caption{Functional contributions and update distributions across architectural components. 
    (a) Effect of embedding layer replacement on MATH500. 
    (b) Layer-wise update norms (bars, left axis) for RL/OPD-trained Qwen3-8B-Base models, and corresponding \textbf{RL} reasoning accuracy after sliding-window intervention (line, right axis) on MATH500.}
    \label{appendix3}
\end{figure}

\begin{figure}[t]
    \centering
    \includegraphics[width=0.8\textwidth]{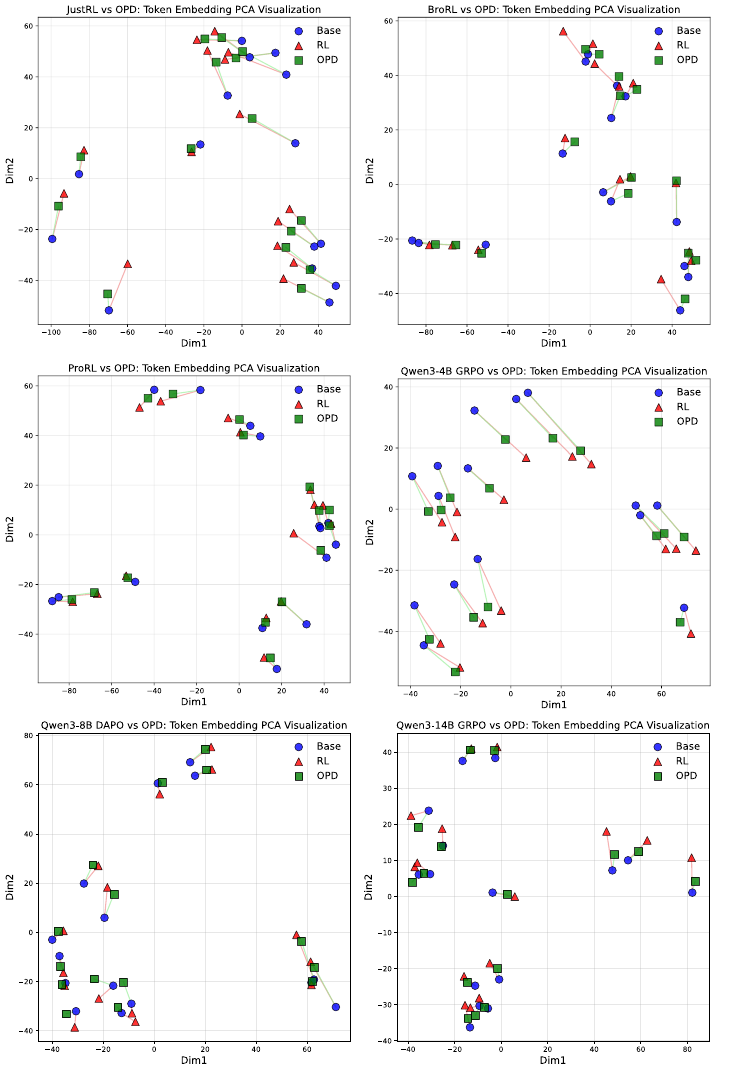}
    \caption{t-SNE visualization of token embeddings from the Base, RL, and OPD models. The red and green lines indicate the shifts from Base to RL and from Base to OPD, respectively.}
    \label{appendix4}
\end{figure}

Finally, we validate the component-level properties identified in the main text under a broader range of datasets and algorithmic settings. These properties include the negligible contribution of embedding layers, the functional dominance of middle-layer MLPs, and the consistent redundancy suppression pattern across architectural components. As shown in Figure~\ref{appendix3}, the results consistently support these observations, further reinforcing that Property 1 reflects an intrinsic and stable characteristic of OPD's parameter update dynamics, rather than an artifact of specific experimental conditions.

\clearpage
\subsection{Detailed Setup of Sliding-Window Intervention Analysis}
\label{Detailed Setup of Sliding-Window Intervention Analysis}

This section provides a formal description of the sliding-window intervention analysis used in Section~\ref{Locating the Residual Updates}. The goal of this analysis is to localize the contribution of parameter updates across different layers and modules \citep{cai2024locatingmitigatinggenderbias,cai2025predictability}, and to examine whether redundant updates in reinforcement learning (RL) are primarily concentrated in functionally non-critical regions.

The core idea of this method is to inject parameter updates into localized regions of the network and measure the resulting performance change \citep{meng2023locatingeditingfactualassociations, vig2020investigating}. Compared to full-model replacement, this localized intervention allows us to isolate the marginal functional contribution of updates at different depths, thereby enabling a fine-grained characterization of the relationship between update location and functional impact.

We consider a Transformer model with \(L\) layers, where each layer consists of two core modules: Attention and MLP. Let \(\Delta W_{\text{RL/OPD}}^{(i,\text{Attn})}\) and \(\Delta W_{\text{RL/OPD}}^{(i,\text{MLP})}\) denote the parameter updates of the Attention and MLP modules at layer \(i\), respectively.

For a target layer \(l\), we define the sliding window as:
\begin{equation}
\mathcal{W}_l = \left\{ i \in \mathbb{Z} \;\middle|\; \max(1,\, l-8) \le i \le \min(L,\, l+8) \right\}.
\end{equation}

The window is centered at layer \(l\) and extends 8 layers to both sides, resulting in a maximum width of 17 layers. Near the model boundaries, the window is truncated accordingly. This design balances locality and stability: by covering neighboring layers, it mitigates the high variance associated with single-layer interventions while preserving spatial resolution.

To isolate the independent contributions of MLP and Attention modules, we construct two types of intervened models. In each setting, only the parameters of the specified module within the sliding window are replaced, while all other parameters are fixed to those of the base model.

\textbf{MLP Intervention:}
\begin{equation}
W_{\text{MLP}, l}^{\text{(interv)}} =
\begin{cases}
    W_{\text{Base}}^{(i,\text{MLP})} + \Delta W_{\text{RL/OPD}}^{(i,\text{MLP})}, & i \in \mathcal{W}_l \\
    W_{\text{Base}}^{(i,\text{MLP})}, & i \notin \mathcal{W}_l.
\end{cases}
\end{equation}
All Attention parameters are fixed to \(W_{\text{Base}}^{(i,\text{Attn})}\).

\textbf{Attention Intervention:}
\begin{equation}
W_{\text{Attn}, l}^{\text{(interv)}} =
\begin{cases}
    W_{\text{Base}}^{(i,\text{Attn})} + \Delta W_{\text{RL/OPD}}^{(i,\text{Attn})}, & i \in \mathcal{W}_l \\
    W_{\text{Base}}^{(i,\text{Attn})}, & i \notin \mathcal{W}_l.
\end{cases}
\end{equation}
All MLP parameters are fixed to \(W_{\text{Base}}^{(i,\text{MLP})}\).

This intervention strategy effectively constructs a \emph{local update injection -- global performance response} analysis framework, allowing us to attribute overall performance changes to specific layers and modules, and thereby reveal the functional distribution of parameter updates across the network.

In practice, we iterate over all valid window centers \(l = 1, 2, \dots, L-8\), construct the two types of intervened models for each \(l\), and evaluate their accuracy on MATH500~\citep{lightman2023lets}. Each intervened model is evaluated using four independent forward passes, and the results are averaged to reduce evaluation noise.


\newpage

\begin{table}[t]
\centering
\caption{Cosine similarity between RL/OPD with Base model token embeddings.}
\setlength{\tabcolsep}{4pt}
\label{embed_similarity}
\begin{tabular}{lcc}
\toprule
\textbf{Base Model} & \textbf{Model} & \textbf{Cosine Sim.} \\
\midrule
\multirow{2}{*}{Qwen2.5-1.5B-Deepseek} & JustRL \citep{he2025justrlscaling15bllm} & 0.9156 \\
                                     & OPD & 0.9412 \\
\midrule
\multirow{2}{*}{Qwen2.5-1.5B-Deepseek} & BroRL \citep{Hu2025BroRLSR} & 0.9078 \\
                  & OPD & 0.9371 \\
\midrule
\multirow{2}{*}{Qwen2.5-1.5B-Deepseek} & ProRL \citep{Liu2025ProRLPR} & 0.9287 \\
                  & OPD & 0.9514 \\
\midrule
\multirow{2}{*}{Qwen3-4B} & Qwen-4B-GRPO \citep{Yang2026LearningBT} & 0.9672 \\
                          & OPD & 0.9851 \\
\midrule
\multirow{2}{*}{Qwen3-8B-Base} & Qwen3-8B-DAPO \citep{cai2025predictability} & 0.9421 \\
                              & OPD & 0.9752 \\
\midrule
\multirow{2}{*}{Qwen3-14B-Base} & Qwen3-14B-Base-DAPO & 0.8961 \\
                                & OPD & 0.9512 \\
\bottomrule
\end{tabular}
\end{table}

\newpage
\clearpage
\section{Property 2 Additional Experiment}
\label{Property 2 Additional Experiment}

\subsection{Geometric Metrics for Parameter Update Matrix}
\label{Geometric Metrics for Parameter Update Matrix}

In this section, we provide formal definitions of four complementary metrics used to characterize the geometric structure of the parameter update matrix $\Delta W \in \mathbb{R}^{m \times n}$. Let the singular value decomposition (SVD) of $\Delta W$ be:
\begin{equation}
\Delta W = U \Sigma V^\top, \quad \Sigma = \operatorname{diag}(\sigma_1, \sigma_2, \ldots, \sigma_r),
\end{equation}
where $r = \operatorname{rank}(\Delta W)$ and $\sigma_1 \ge \sigma_2 \ge \cdots \ge \sigma_r > 0$ are the singular values sorted in descending order.

\paragraph{Spectral Norm~\citep{MATHIAS1990269}.}
The spectral norm is defined as the largest singular value $\sigma_1.$. This metric captures the magnitude of the update along the dominant direction in parameter space, corresponding to the maximum amplification induced by $\Delta W$ on any input vector.

\paragraph{Spectral-to-Frobenius Norm Ratio~\citep{ALNATOOR2024e30056}.}
The spectral-to-Frobenius norm ratio is defined as:
\begin{equation}
\rho = \frac{\sigma_1}{\sqrt{\sum_{j=1}^r \sigma_j^2}}.
\end{equation}
This ratio quantifies the dominance of the leading singular direction. A value of $\rho$ close to $1$ indicates that the update is highly concentrated along a single direction, whereas smaller values suggest that the update energy is distributed across multiple directions.

\paragraph{Effective Rank~\citep{item_f3c74b8f1cad43ed869604b318d58703}.}
The effective rank, also referred to as the spectral entropy rank, is defined as:
\begin{equation}
\mathrm{rank}_{\mathrm{eff}} = \exp\left(-\sum_{i=1}^r \bar{\sigma}_i \log \bar{\sigma}_i\right),
\end{equation}
where $\bar{\sigma}_i = \sigma_i / \sum_{j=1}^r \sigma_j$ denotes the normalized singular values. This metric measures the entropy of the singular value spectrum. A smaller effective rank indicates rapid spectral decay and concentration of update energy in a low-dimensional subspace, while a larger effective rank implies a more diffuse distribution.

\paragraph{Top-1\% Subspace Norm Ratio~\citep{cai2025predictability}.}
Let $k = \lceil r / 100 \rceil$ denote the number of singular components corresponding to the Top $1\%$ of the spectrum. We construct the rank-$k$ approximation of $\Delta W$ using these leading components:
\begin{equation}
\Delta W_k = U_{:,1:k} \Sigma_{1:k,1:k} V_{:,1:k}^\top.
\end{equation}
The Top-$1\%$ subspace norm ratio is defined as:
\begin{equation}
R_{\text{Top-1\%}} = \frac{\|\Delta W_k\|_F}{\|\Delta W\|_F}
= \sqrt{\frac{\sum_{i=1}^k \sigma_i^2}{\sum_{j=1}^r \sigma_j^2}}.
\end{equation}
This metric quantifies the fraction of the total update energy captured by the Top $1\%$ of singular directions. A value close to $1$ indicates that the update is effectively confined to an extremely low-dimensional subspace. For each model, we report the average values of the computed metrics across all MLP and attention matrices.

\clearpage

\begin{figure}[h] 
    \centering
    \includegraphics[width=0.9\textwidth]{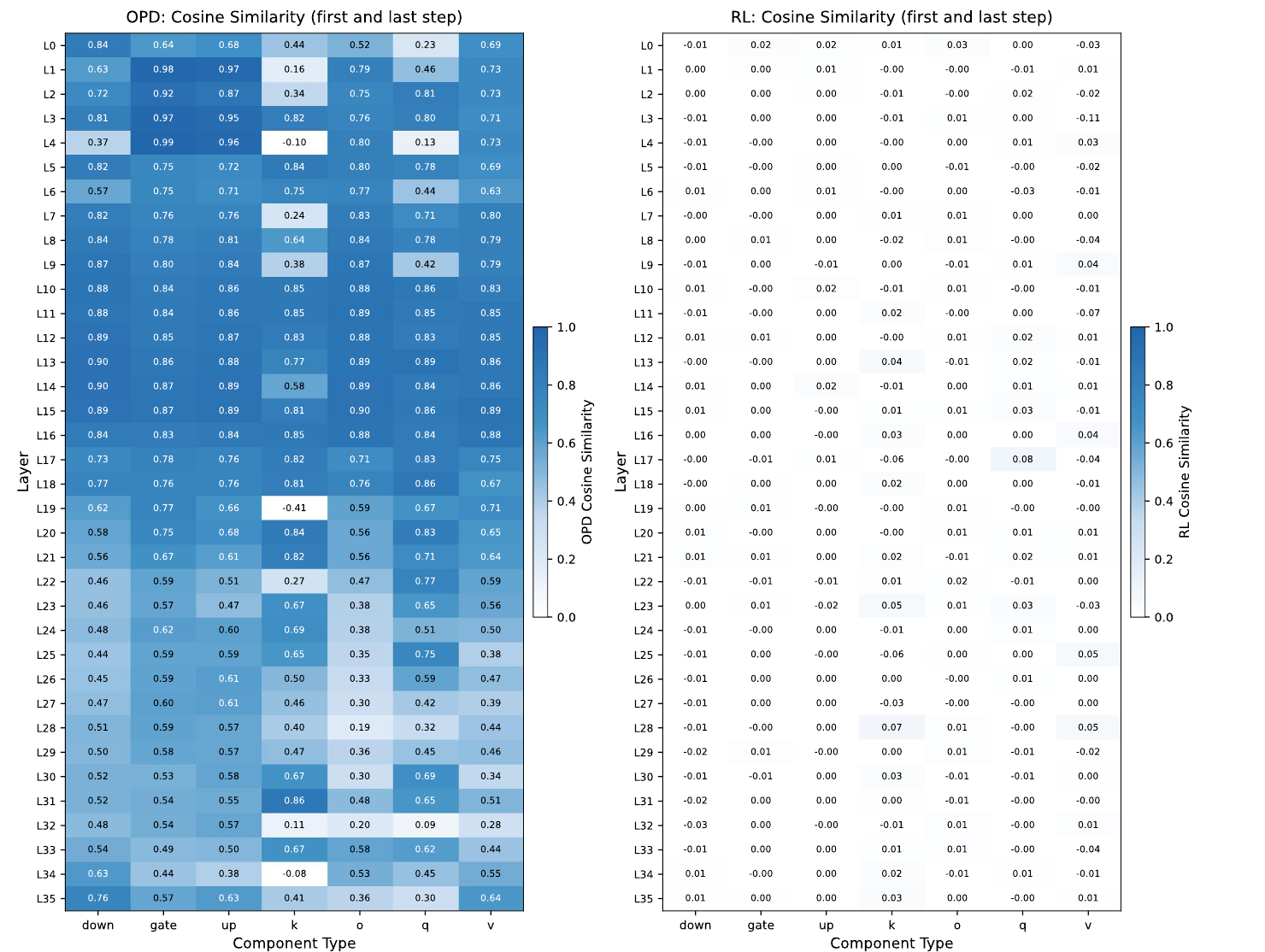} 
    \caption{Heatmap of cosine-similarity of $\mathcal{U}_1$ at the first and last steps for each component trained under OPD and RL.}
    \label{cosappendix}
\end{figure}

\subsection{Cosine Similarity Analysis of Subspaces}
\label{Cosine Similarity Analysis of Subspace}

This section provides additional empirical evidence for Property 2 (Early Low-Rank Lock-in) by analyzing the directional stability of dominant update subspaces during training. We focus on how the principal subspaces evolve from the early training stage to the final converged checkpoint, thereby characterizing the subspace-level convergence behavior of different training methods.

To this end, we perform singular value decomposition (SVD) on the parameter update matrix and analyze the dominant subspaces spanned by its leading singular vectors. Specifically, we consider the Rank-1 subspace \(\mathcal{U}_1\), which corresponds to the strongest singular direction and captures the primary low-dimensional structure of update energy. We compute the cosine similarity between early-stage and final-stage subspaces to measure the degree of directional lock-in during training. The results are shown in Figure~\ref{cosappendix}.

\textbf{RL exhibits unstable dominant subspace evolution.}
During RL training, the cosine similarity between early-stage and final-stage subspaces remains consistently low across modules. This indicates that RL does not establish update directions aligned with the final checkpoint at the early stage. Instead, its dominant subspaces undergo substantial changes throughout training, suggesting that RL requires continuous exploration and correction before gradually converging to a stable configuration.

\textbf{OPD exhibits early alignment of dominant subspaces.}
In contrast, OPD shows substantially higher subspace consistency across most modules. In particular, intermediate layers exhibit especially strong early alignment, with cosine similarity reaching up to 0.9. These results indicate that OPD identifies stable dominant update directions early in training, while subsequent optimization mainly amplifies the update magnitude along these directions rather than repeatedly searching for new directions.

This observation provides further support for Property 1 from a representational geometry perspective. As Property 1 indicates, OPD suppresses functionally redundant updates and concentrates parameter changes within reasoning-critical intermediate modules. The present subspace analysis elucidates the mechanistic basis for such compact updates: in these modules, the dominant update subspaces stabilize early during training, enabling OPD to amplify updates along these consistent directions while minimizing redundant parameter movement. Consequently, OPD achieves substantial performance improvements with high parameter efficiency, as the optimization primarily reinforces already stable, task-relevant directions rather than exploring unnecessary or redundant dimensions.

\begin{figure}[h] 
    \centering
    \includegraphics[width=0.9\textwidth]{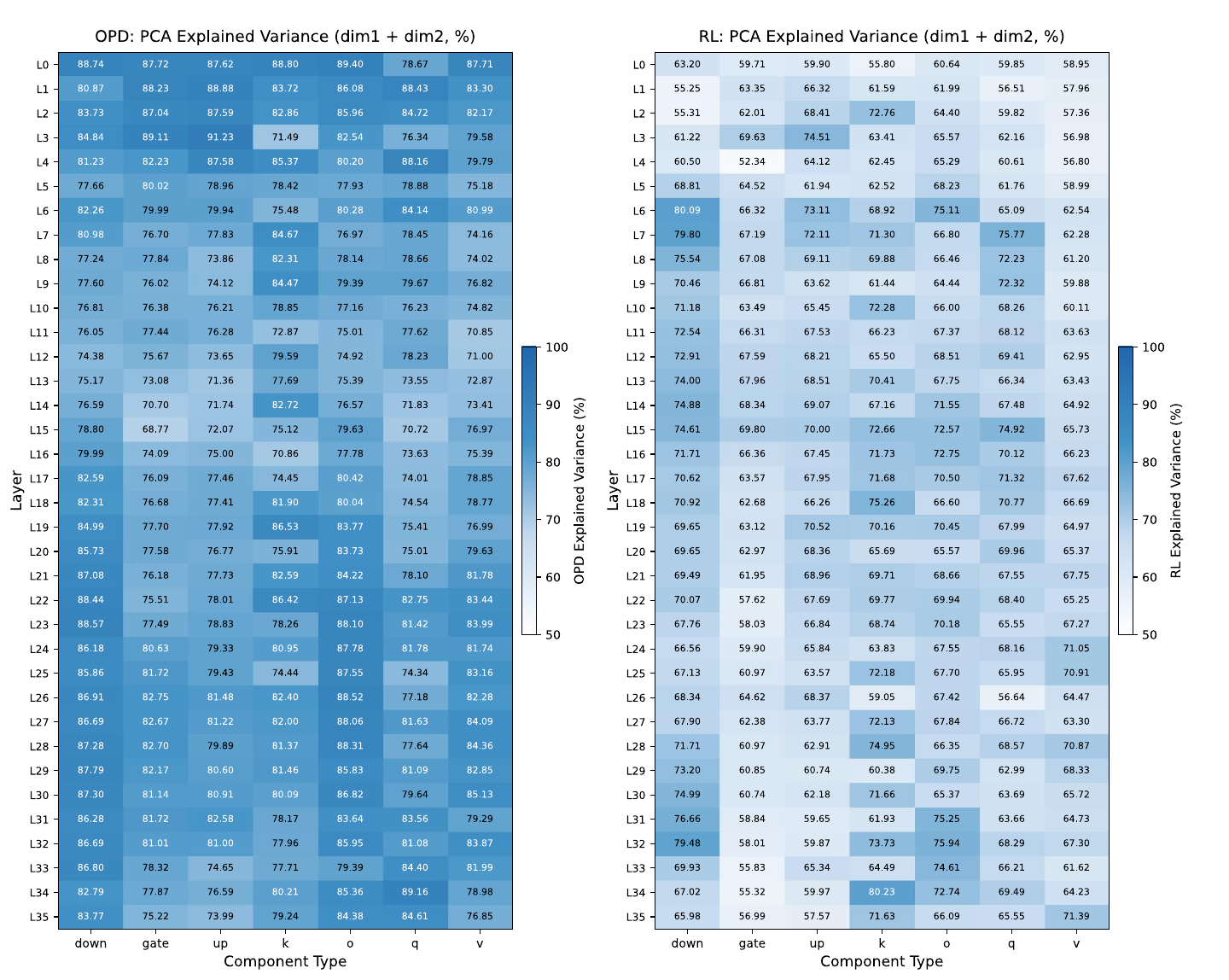} 
    \caption{Heatmap of $\mathcal{U}_1$ trajectory under OPD and RL, along with variance explained by the first two dimensions after PCA.}
    \label{PCAappendix}
\end{figure}

\subsection{Trajectory Evolution of Subspaces}

\textbf{Trajectory Visualization.}
Beyond similarity analysis, we further investigate the temporal evolution of dominant subspaces during training by visualizing the trajectories of Rank-1 subspaces \(\mathcal{U}_1\) across different modules. Specifically, we apply t-SNE dimensionality reduction~\citep{shi2021visualizingdatausinggtsne} to representations from different training checkpoints, with results shown in Figures~\ref{tsne_grid_mlp_down_proj}-\ref{tsne_grid_self_attn_o_proj (1)}.

We observe that OPD exhibits markedly more concentrated trajectory patterns: its evolution is confined to a narrower region in the projected space and follows a smoother, near-linear path. In contrast, RL trajectories are significantly more dispersed and irregular. This suggests that OPD induces stronger directional stability during representation evolution, resulting in a more structured and predictable optimization trajectory.

\textbf{Quantitative Characterization via PCA.}
To quantify this phenomenon, we perform PCA~\citep{Eckart1936TheAO} on representations from different training checkpoints. For each module, we collect the checkpoint-wise representation vectors and form a trajectory matrix \(X \in \mathbb{R}^{T \times d}\), where \(T\) denotes the number of checkpoints and \(d\) is the representation dimension. After centering \(X\), PCA decomposes the covariance matrix and obtains eigenvalues \(\lambda_1 \geq \lambda_2 \geq \cdots \geq \lambda_d\). We then compute the cumulative variance explained by the first two principal components as
\begin{equation}
\mathrm{EVR}_{0:2}
=
\frac{\lambda_1 + \lambda_2}{\sum_{i=1}^{d} \lambda_i}.
\end{equation}
This quantity measures how much of the trajectory variation across training checkpoints can be captured by a two-dimensional principal subspace. A higher value indicates that the trajectory is more concentrated and lower-dimensional, whereas a lower value suggests that the evolution is more dispersed across multiple directions. The results are summarized in Figure~\ref{PCAappendix}.

Overall, OPD consistently achieves substantially higher \(\mathrm{EVR}_{0:2}\) than RL. This indicates that the OPD representations are more strongly concentrated within a low-dimensional and compact subspace during training. In contrast, RL representations distribute their variation across a broader set of directions, reflecting greater redundancy and less structured trajectory evolution.

\textbf{Mechanistic Interpretation.}
Overall, these observations provide a unified geometric and information-theoretic perspective on the behaviors described in Property 1 and Property 2. Specifically, during training, the update dynamics are not evenly distributed across the high-dimensional parameter space but are highly concentrated along a few dominant directions forming a low-dimensional subspace. From an information-theoretic standpoint, this concentration acts as a form of implicit compression, enhancing parameter utilization efficiency (Property 1) while facilitating early stabilization of update directions (Property 2).

From the perspective of optimization geometry, this concentration reflects an implicit low-rank bias: under dense teacher supervision, OPD preferentially updates along a small number of stable and effective directions rather than exploring the high-dimensional parameter space indiscriminately. As a result, the parameter evolution exhibits a highly structured pattern, with both the direction and support of updates tightly constrained, yielding compact and stable trajectory evolution.

\newpage

\begin{figure}[h] 
    \centering
    \includegraphics[width=0.9\textwidth]{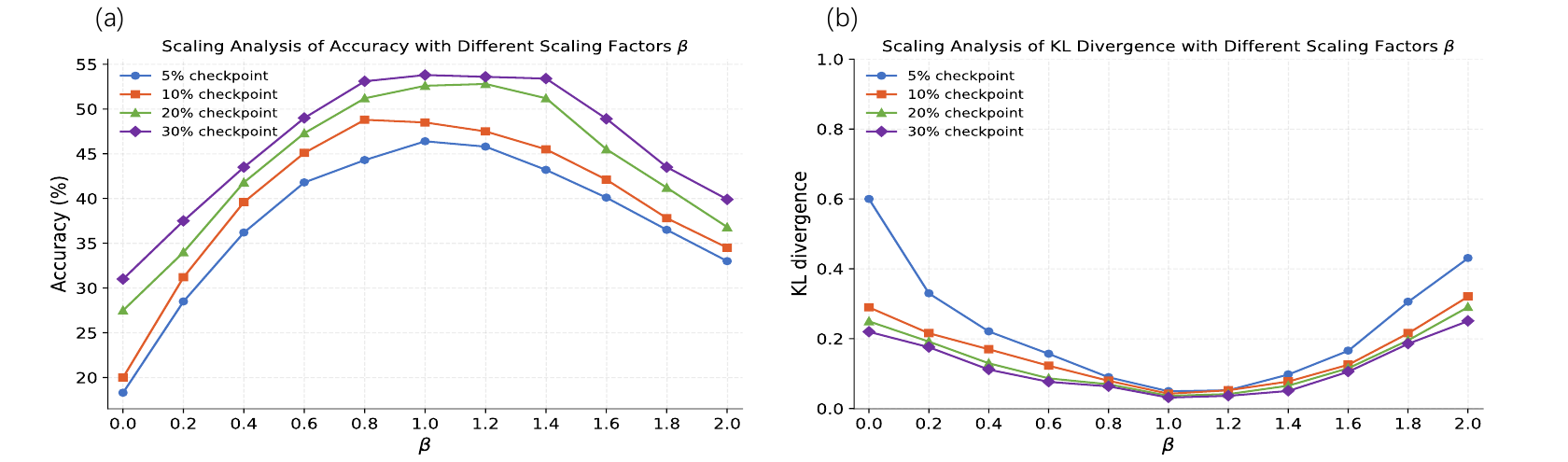} 
    \caption{Scaling analysis of (a) accuracy and (b) KL divergence across different training checkpoints, with optimal performance achieved in the range $0.8 \leq \beta \leq 1.2$.}
    \label{scaleappendix}
\end{figure}

\subsection{Scaling Effects on Accuracy and Distribution Alignment}
\label{Scaling Effects on Accuracy and Distribution Alignment}
This subsection aims to further validate and complement the findings in Section \ref{section3} Figure \ref{fig4}, focusing on the relationship between the magnitude of early updates and model performance.

\paragraph{Effect of Scaling Magnitude on Performance.} To analyze the effect of scaling early checkpoint updates on model performance, we define the updated parameters after scaling as:

\begin{equation}
\Delta W_{\text{scaled}} = \Delta W_{\text{early}} + \underbrace{\Delta W_{\text{early}} \times \frac{\beta \cdot (\|\Delta W_{\text{final}}\|_F - \|\Delta W_{\text{early}}\|_F)}{\|\Delta W_{\text{early}}\|_F}}_{\text{extra update}}.
\end{equation}

Here, \(\beta\) is the scaling coefficient. When \(\beta = 0\), \(\Delta W_{\text{scaled}} = \Delta W_{\text{early}}\), i.e., no extra update is added. When \(\beta = 1\), \(\|\Delta W_{\text{scaled}}\|_F = \|\Delta W_{\text{final}}\|_F\), i.e., the magnitude of the scaled update matches that of the final update.

As shown in Figure~\ref{scaleappendix} (a), increasing $\beta$ from 0 progressively improves model performance. When $\beta \approx 0.8$, the performance gain begins to plateau; when $\beta$ exceeds a large value (approximately $1.2$), performance starts to degrade. This trend provides three key insights: (i) the early checkpoint already captures a principal subspace aligned with the final solution, as evidenced by performance gains from moderate scaling; (ii) the plateau around $\beta \approx 0.8$ reflects inherent representational limits of the early subspace, indicating that further amplification cannot fully bridge the gap without additional training; (iii) excessive scaling leads to performance decline, suggesting that extra norm amplifies noise or irrelevant components, harming task performance.

\paragraph{Alignment with Teacher Distribution.} 
To further understand these trends, we measure the KL divergence between the student’s outputs and the teacher’s distribution. Figure~\ref{scaleappendix} (b) shows that KL divergence decreases monotonically with increasing $\beta$, stabilizes over the intermediate range corresponding to the performance plateau, and rises again for $\beta > 1.2$. These trends mirror the accuracy results: initially, monotonic KL reduction coincides with steady accuracy improvement, indicating that closer approximation to the teacher distribution directly drives task performance. Within the optimal range ($\beta \approx 0.8$--$1.2$), KL divergence remains low and accuracy saturates, demonstrating strong student-teacher distribution alignment.

This phenomenon can be interpreted from two complementary perspectives. First, from a causal inference viewpoint, KL reduction—i.e., more precise alignment with the teacher’s behavioral distribution—directly drives improvements in task accuracy. Second, from the perspective of representation subspace geometry, the reduction in KL following scaling reveals that the early update directions already capture the dominant structure of the teacher’s distribution. While the early subspace norm may initially be insufficient, its directions are largely aligned with the final converged solution. Appropriate scaling partially unlocks the representational capacity encoded in this subspace, thereby reducing the distributional gap between student and teacher.

\paragraph{Illustrative example of scaling-induced reasoning improvement.}
Then, we provide a concrete example to illustrate the differences in text representations between the early checkpoint and the teacher model. On the next page, we compare the generated responses of the early checkpoint before and after scaling. Specifically, when we scale the norm of the early checkpoint to match that of the final model, the quality of its generated responses improves significantly compared to the unscaled version. Further analysis reveals that the scaled responses exhibit a noticeable increase in the number of reasoning steps, with each step becoming more fine-grained. The model demonstrates richer intermediate reasoning processes and clearer logical progression, rather than jumping directly to results. This change reflects reasoning habits that are more similar to those of the teacher model, indicating that appropriate norm scaling can activate the reasoning structures already encoded in the early subspace, making the student's generation behavior more akin to the teacher's in terms of reasoning depth and logical coherence.

\subsection{A Local Geometric View of OPD Dynamics}
\label{A Linearized View of OPD Dynamics}

In this subsection, we provide a local geometric analysis to explain why On-Policy Distillation (OPD) naturally induces low-rank and early-locked update directions, and how this differs from the update dynamics of reinforcement learning (RL). By linearizing the student model around the base model, we reveal how the structure of the OPD objective gives rise to the empirical phenomena observed in the main text.

\paragraph{Setup and Linearization.}
Let a token context be denoted by \(c = (x, y_{<t})\), where \(x\) is the input prompt and \(y_{<t}\) are previously generated tokens.  
Define:
\begin{itemize}
    \item \(z_{\theta}(c) \in \mathbb{R}^{V}\): logits of the student model with parameters \(\theta\) (vocabulary size \(V\)).
    \item \(z^{\star}(c) \in \mathbb{R}^{V}\): logits of a fixed teacher model.
    \item \(\theta_0\): parameters of the base model (initialization for both RL and OPD training).
    \item \(\Delta\theta = \theta - \theta_0\): parameter displacement.
\end{itemize}

Expand \(z_{\theta}(c)\) around \(\theta_0\) to first order:
\begin{equation}
z_{\theta}(c) = z_{\theta_0}(c) + \underbrace{\frac{\partial z_{\theta}(c)}{\partial \theta}\bigg|_{\theta=\theta_0}}_{=: J_c} \Delta\theta + O(\|\Delta\theta\|^2).
\end{equation}
Here \(J_c \in \mathbb{R}^{V \times \dim(\theta)}\) is the Jacobian matrix of the logits with respect to the parameters.  
For sufficiently small step sizes and early training, \(\|\Delta\theta\|\) is small, and we neglect the higher-order terms:
\begin{equation}
z_{\theta}(c) \approx z_{0}(c) + J_c \Delta\theta, \qquad \text{where } z_{0}(c):=z_{\theta_0}(c).
\end{equation}

Define the \emph{teacher-student logit residual at the base model}:
\begin{equation}
r_c := z^{\star}(c) - z_{0}(c).
\end{equation}
Then the logit discrepancy becomes:
\begin{equation}
z_{\theta}(c) - z^{\star}(c) \approx J_c \Delta\theta - r_c.
\end{equation}

\paragraph{Local Quadratic Approximation of the OPD Objective.}
The OPD objective minimizes the reverse KL divergence between the student and the teacher on on-policy samples:
\begin{equation}
\mathcal{L}_{\mathrm{OPD}}(\theta) = \mathbb{E}_{x\sim\mathcal{D},\; y\sim\pi_{\theta}(\cdot|x)}\left[ D_{\mathrm{KL}}\big(\pi_{\theta}(\cdot|x,y_{<t})\;\|\; \pi^{\star}(\cdot|x,y_{<t})\big) \right].
\end{equation}
For a fixed context \(c\), denote:
\begin{equation}
p_{\theta}(\cdot|c) = \mathrm{softmax}(z_{\theta}(c)), \qquad p^{\star}(\cdot|c) = \mathrm{softmax}(z^{\star}(c)).
\end{equation}

When the two distributions are close, the KL divergence admits a second-order Taylor expansion in the logit space. Let \(f(z) = D_{\mathrm{KL}}(p_z\|p^{\star})\) where \(p_z = \mathrm{softmax}(z)\). Then:
\begin{equation}
f(z) \approx f(z^{\star}) + \underbrace{\nabla f(z^{\star})^{\top}(z-z^{\star})}_{=0} + \frac{1}{2}(z-z^{\star})^{\top} \nabla^2 f(z^{\star}) (z-z^{\star}),
\end{equation}
because the first derivative vanishes at \(z=z^{\star}\) (minimum). The Hessian of the reverse KL at the teacher point is the Fisher information matrix of the student distribution:
\begin{equation}
\nabla^2 f(z^{\star}) = \mathrm{Diag}(p^{\star}) - p^{\star}{p^{\star}}^{\top} =: F_c^{\star}.
\end{equation}
Thus, for \(z\) near \(z^{\star}\):
\begin{equation}
D_{\mathrm{KL}}(p_z\|p^{\star}) \approx \frac{1}{2} (z - z^{\star})^{\top} F_c^{\star} (z - z^{\star}).
\end{equation}

However, in our local analysis we linearize around \(\theta_0\), so the student logits \(z_{\theta}(c)\) are close to \(z_0(c)\), not necessarily close to \(z^{\star}(c)\). To obtain a quadratic form in \(\Delta\theta\), we may evaluate the Fisher matrix at a convenient distribution, typically the base model distribution \(p_0(c) = \mathrm{softmax}(z_0(c))\). This yields an approximation that is consistent when \(z_{\theta} \approx z_0\) and the teacher is not too far from the base model. Define:
\begin{equation}
F_c := \mathrm{Diag}(p_0(c)) - p_0(c)p_0(c)^{\top}.
\end{equation}
Then we approximate:
\begin{equation}
D_{\mathrm{KL}}(p_{\theta}\|p^{\star}) \approx \frac{1}{2} (z_{\theta}(c) - z^{\star}(c))^{\top} F_c (z_{\theta}(c) - z^{\star}(c))
= \frac{1}{2} (J_c\Delta\theta - r_c)^{\top} F_c (J_c\Delta\theta - r_c).
\end{equation}

If the teacher and base model are already reasonably aligned (a common scenario in distillation), then \(z^{\star} \approx z_0\) and \(F_c \approx F_c^{\star}\). Even if not, the quadratic form still provides a local approximation of the KL divergence up to an additive constant, because:
\begin{equation}
D_{\mathrm{KL}}(p_{\theta}\|p^{\star}) = D_{\mathrm{KL}}(p_0\|p^{\star}) + \nabla_{\theta}D_{\mathrm{KL}}(p_{\theta}\|p^{\star})|_{\theta_0} \Delta\theta + \frac{1}{2}\Delta\theta^{\top} H \Delta\theta + \cdots,
\end{equation}
and the Hessian at \(\theta_0\) involves \(J_c^{\top} F_c^{\star} J_c\). Evaluating \(F_c\) at \(p_0\) is a standard simplification in the neural tangent kernel literature and preserves the correct second-order structure when \(\|z^{\star}-z_0\|\) is small.

\paragraph{Local Expected Objective and Gradient.}
Taking expectation over the on-policy contexts \(c\) which, to first order, can be approximated by the base model's distribution, we obtain:
\begin{equation}
\mathcal{L}_{\mathrm{OPD}}(\Delta\theta) \approx \frac{1}{2} \mathbb{E}_{c}\big[ (J_c\Delta\theta - r_c)^{\top} F_c (J_c\Delta\theta - r_c) \big].
\end{equation}
Expanding the quadratic:
\begin{equation}
\mathcal{L}_{\mathrm{OPD}}(\Delta\theta) \approx 
\frac{1}{2} \Delta\theta^{\top} \underbrace{\mathbb{E}_c[J_c^{\top} F_c J_c]}_{=: A} \Delta\theta
- \Delta\theta^{\top} \underbrace{\mathbb{E}_c[J_c^{\top} F_c r_c]}_{=: b}
+ \frac{1}{2} \mathbb{E}_c[r_c^{\top} F_c r_c].
\end{equation}
The last term is constant with respect to \(\Delta\theta\). Therefore, the local objective is a convex quadratic:
\begin{equation}
\mathcal{L}_{\mathrm{OPD}}(\Delta\theta) = \frac{1}{2} \Delta\theta^{\top} A \Delta\theta - b^{\top} \Delta\theta + \text{const}.
\end{equation}
The gradient with respect to \(\Delta\theta\) is:
\begin{equation}
g(\Delta\theta) := \nabla_{\Delta\theta} \mathcal{L}_{\mathrm{OPD}} = A \Delta\theta - b.
\end{equation}

\paragraph{Gradient Descent Dynamics and Closed-Form Solution.}
Consider gradient descent on \(\Delta\theta\) with fixed step size \(\eta > 0\):
\begin{equation}
\Delta\theta_{s+1} = \Delta\theta_s - \eta g(\Delta\theta_s) = \Delta\theta_s - \eta (A\Delta\theta_s - b) = (I - \eta A)\Delta\theta_s + \eta b.
\end{equation}
Starting from \(\Delta\theta_0 = 0\) (initialization at the base model), we unroll the recursion:
\begin{align}
\Delta\theta_1 &= \eta b, \\
\Delta\theta_2 &= (I - \eta A)\eta b + \eta b = \eta[I + (I - \eta A)] b, \\
\Delta\theta_s &= \eta \sum_{j=0}^{s-1} (I - \eta A)^{j} b.
\end{align}
This is a geometric series of matrices. Assume \(A\) is symmetric positive semidefinite (it is a Gram matrix of \(J_c^{\top}F_c^{1/2}\)). Choose \(\eta\) such that \(0<\eta < 2/\lambda_{\max}(A)\) to ensure convergence. Then \(I - \eta A\) has spectral radius less than 1, and the series converges to:
\begin{equation}
\Delta\theta_{\infty} = \eta (I - (I - \eta A))^{-1} b = A^{-1} b,
\end{equation}
where \(A^{-1}\) denotes the pseudo-inverse on the support of \(A\). The finite-sum formula can be expressed in closed form:
\begin{equation}
\Delta\theta_s = \big[ I - (I - \eta A)^s \big] A^{-1} b.
\end{equation}
This is verified by factoring:
\begin{equation}
\sum_{j=0}^{s-1} (I - \eta A)^j = (I - (I - \eta A)^s)(I - (I - \eta A))^{-1} = (I - (I - \eta A)^s)(\eta A)^{-1}.
\end{equation}
Multiplying by \(\eta b\) gives the result.

\paragraph{Spectral Decomposition and Directional Dynamics.}
Let \(A = U \Lambda U^{\top}\) be the eigen-decomposition with \(\Lambda = \mathrm{diag}(\lambda_1,\lambda_2,\dots,\lambda_d)\) and \(\lambda_1 \ge \lambda_2 \ge \dots \ge \lambda_d \ge 0\).  
Let \(b = U \beta\) with \(\beta_i = \langle b, u_i \rangle\). Since \(A^{-1} = U \Lambda^{-1} U^{\top}\) (pseudo-inverse), we have:
\begin{equation}
A^{-1}b = \sum_{i: \lambda_i>0} \frac{\beta_i}{\lambda_i} u_i.
\end{equation}
Also, \((I - \eta A)^s = U (I - \eta \Lambda)^s U^{\top}\). Therefore:
\begin{equation}
\Delta\theta_s = U \big[ I - (I - \eta \Lambda)^s \big] \Lambda^{-1} \beta = \sum_{i: \lambda_i>0} \frac{1 - (1 - \eta \lambda_i)^s}{\lambda_i} \beta_i u_i.
\end{equation}

The above expression reveals the directional dynamics. For each eigen-direction \(u_i\), the contribution starts at zero and asymptotically approaches \(\beta_i/\lambda_i\). The factor \(1-(1-\eta\lambda_i)^s\) grows more rapidly when the curvature \(\lambda_i\) is larger, meaning that directions with high sensitivity of the logits to parameter changes saturate early. Consequently, if the projection \(\beta_i\) vanishes for many directions, the effective update remains confined to a low‑dimensional subspace throughout training.

\paragraph{A Sufficient Condition for Early Low-Rank Lock-in.}
Define the top-\(k\) eigenspace of \(A\) as
\[
U_k=\mathrm{span}\{u_1,\dots,u_k\},
\]
and let \(P_{U_k}\) be the orthogonal projector onto this subspace. We assume that the driving term \(b\) is concentrated in \(U_k\) up to a small residual:
\begin{equation}
\|P_{U_k^\perp}b\|\leq \epsilon\|b\|,
\qquad
\epsilon\ll 1.
\end{equation}
Equivalently, we decompose
\[
b=b_{\parallel}+b_{\perp},
\qquad
b_{\parallel}=P_{U_k}b,
\qquad
b_{\perp}=P_{U_k^\perp}b .
\]
Using the closed-form dynamics, the update can be written as
\begin{equation}
\Delta\theta_s
=
[I-(I-\eta A)^s]A^{-1}b_{\parallel}
+
[I-(I-\eta A)^s]A^{-1}b_{\perp}.
\end{equation}
The first term lies in the dominant eigenspace \(U_k\), while the second term corresponds to the tail contribution from \(U_k^\perp\). Rather than assuming that \(A^{-1}\) is norm-reducing on the orthogonal complement, we bound this tail term through the spectral response of the finite-step dynamics. Specifically,
\begin{equation}
\left\|
[I-(I-\eta A)^s]A^{-1}b_{\perp}
\right\|
\leq
\rho_{\perp}(s)\|b_{\perp}\|,
\end{equation}
where
\begin{equation}
\rho_{\perp}(s)
=
\max_{i>k,\lambda_i>0}
\frac{\left|1-(1-\eta\lambda_i)^s\right|}{\lambda_i}.
\end{equation}
Combining this with the concentration assumption gives
\begin{equation}
\left\|
\Delta\theta_s
-
[I-(I-\eta A)^s]A^{-1}b_{\parallel}
\right\|
\leq
\rho_{\perp}(s)\epsilon\|b\|.
\end{equation}

Thus, when the projected residual \(b\) is highly concentrated in the top-\(k\) eigenspace, the tail contribution remains small during the finite training horizon. If, in addition, there is a clear spectral gap,
\begin{equation}
\lambda_k \gg \lambda_{k+1},
\end{equation}
then the dominant directions in \(U_k\) are activated and saturated earlier than the tail directions. This provides a geometric explanation for \textbf{Property 2 (Early Low-Rank Lock-in)}: the optimization path is largely confined to a low-dimensional subspace that is identified in the early stage of training, while subsequent optimization mainly increases the magnitude within this subspace rather than exploring substantially new directions.

\emph{Why is \(b\) low-rank in practice?}
Recall that
\begin{equation}
b=\mathbb{E}_c[J_c^{\top}F_cr_c].
\end{equation}
The residual
\[
r_c=z^{\star}(c)-z_0(c)
\]
is the teacher-base logit difference. In distillation, the teacher often refines the student by sharpening probabilities on a relatively small set of functionally important token positions, such as key reasoning tokens, intermediate reasoning steps, answer tokens, or formatting tokens \citep{xu2026tip}. Hence, \(r_c\) is often sparse or low-dimensional in its effective support. The Fisher matrix \(F_c\) further reweights these residual directions according to the local geometry of the output distribution. Although \(J_c\) itself can be high-rank, the composition
\[
J_c^{\top}F_cr_c
\]
projects this concentrated residual signal back into parameter space. After averaging over contexts, the resulting driving term \(b\) tends to concentrate on parameter directions that most strongly affect those critical token predictions. This is consistent with the low-rank structure of \(\Delta W\) observed in Section \ref{section3}.

\paragraph{Module-Wise Suppression (Functional Redundancy Avoidance).}
Decompose the parameters into \(M\) modules (e.g., embedding, attention, MLP layers). Write:
\begin{equation}
\Delta\theta = (\Delta\theta_1, \Delta\theta_2, \dots, \Delta\theta_M), \qquad
J_c = [J_{c,1}, J_{c,2}, \dots, J_{c,M}],
\end{equation}
where \(J_{c,m} = \partial z_{\theta}(c)/\partial \theta_m|_{\theta_0}\).  
Then the driving term for module \(m\) is:
\begin{equation}
b_m = \mathbb{E}_c[ J_{c,m}^{\top} F_c r_c ].
\end{equation}
The curvature matrix \(A\) has block structure:
\begin{equation}
A = \begin{pmatrix}
A_{11} & A_{12} & \cdots & A_{1M} \\
A_{21} & A_{22} & \cdots & A_{2M} \\
\vdots  & \vdots  & \ddots & \vdots  \\
A_{M1} & A_{M2} & \cdots & A_{MM}
\end{pmatrix},
\quad A_{mn} = \mathbb{E}_c[J_{c,m}^{\top} F_c J_{c,n}].
\end{equation}

At the local optimum \(\Delta\theta^* = A^{-1}b\) (or the limit of gradient descent), we have:
\begin{equation}
\sum_{n=1}^{M} A_{mn} \Delta\theta_n^* = b_m.
\end{equation}
If the cross-module coupling is weak (i.e., \(A_{mn}\) is small for \(m\ne n\) compared to \(A_{mm}\)), and \(A_{mm}\) is invertible on its support, then:
\begin{equation}
\Delta\theta_m^* \approx A_{mm}^{-1} b_m.
\end{equation}
Thus, if \(b_m \approx 0\) (module \(m\) is weakly coupled with the teacher residual), then \(\Delta\theta_m^* \approx 0\). This provides a mechanism for \textbf{Property 1 (Functional Redundancy Avoidance)}: modules that do not help match the teacher residual receive negligible updates. Empirically, embedding layers and bottom/top transformer layers have small \(b_m\), leading to suppressed updates.

\paragraph{Comparison with Reinforcement Learning Dynamics.}
A standard policy gradient update (e.g., PPO) for a trajectory of length \(T\) is:
\begin{equation}
g_{\mathrm{RL}} = \mathbb{E}_{x\sim\mathcal{D}, y\sim\pi_{\theta}(\cdot|x)}\left[ \sum_{t=1}^{T} A_t \nabla_\theta \log \pi_\theta(y_t|c_t) \right],
\end{equation}
where \(c_t = (x,y_{<t})\) and \(A_t\) is an advantage estimate. Using the logit parameterization:
\begin{equation}
\nabla_\theta \log \pi_\theta(y_t|c_t) = J_{c_t}^{\top} (e_{y_t} - p_\theta(\cdot|c_t)).
\end{equation}
Hence:
\begin{equation}
g_{\mathrm{RL}} = \mathbb{E}\left[ \sum_{t=1}^{T} A_t J_{c_t}^{\top} (e_{y_t} - p_\theta(\cdot|c_t)) \right].
\end{equation}

In contrast, the OPD gradient (local approximation) is:
\begin{equation}
g_{\mathrm{OPD}} = -\nabla_{\Delta\theta}\mathcal{L}_{\mathrm{OPD}} = b - A\Delta\theta.
\end{equation}
At initialization (\(\Delta\theta=0\)), we have \(g_{\mathrm{OPD}}(0) = b\), which is a deterministic (up to sampling) function of the teacher residual. The RL gradient at initialization is:
\begin{equation}
g_{\mathrm{RL}}(0) = \mathbb{E}\left[ \sum_{t} A_t J_{c_t}^{\top} (e_{y_t} - p_0(c_t)) \right].
\end{equation}

The differences between the two paradigms can be summarized in a few key aspects. OPD benefits from dense token‑level supervision through the residual \(r_c\) (filtered by \(F_c\)), whereas RL relies on scalar rewards \(A_t\) that are typically zero for most tokens in sparse reward settings, making RL gradient estimates noisier. Moreover, credit assignment in RL is challenging because \(A_t\) depends on the entire trajectory and future rewards, introducing high variance. In OPD, the per‑token residual provides a more stable learning signal. Finally, the directional structure differs crucially: the OPD driving term \(b\) inherits the low‑rank concentration of \(r_c\), while the RL driving term involves \(e_{y_t} - p_0(c_t)\), a random vector with full support in the vocabulary space, leading to less concentrated and more diffuse updates.

We can approximate the gradient covariance to illustrate the difference. For OPD, the per-sample gradient at initialization is:
\begin{equation}
\hat{g}_{\mathrm{OPD}} = J_c^{\top} F_c r_c,
\end{equation}
with covariance \(\Sigma_{\mathrm{OPD}} = \mathrm{Cov}(\hat{g}_{\mathrm{OPD}})\). For RL, assuming a single-token simplification (or ignoring temporal dependencies), the per-sample gradient is:
\begin{equation}
\hat{g}_{\mathrm{RL}} = A J_c^{\top} (e_y - p_0(c)).
\end{equation}
Its covariance satisfies:
\begin{equation}
\mathrm{Tr}(\Sigma_{\mathrm{RL}}) \approx \mathbb{E}[A^2] \cdot \mathbb{E}[\|J_c^{\top}(e_y-p_0)\|^2] \;\ge\; \sigma_A^2 \cdot \mathbb{E}[\|J_c^{\top}(e_y-p_0)\|^2],
\end{equation}
where \(\sigma_A^2 = \mathrm{Var}(A)\). In sparse-reward settings, \(\sigma_A^2\) can be large because most trajectories receive zero reward except a few. For OPD, the residual \(r_c\) is non-zero for many tokens, leading to lower relative variance. Moreover, the norm \(\|J_c^{\top}(e_y-p_0)\|\) is typically larger in magnitude than \(\|J_c^{\top}F_c r_c\|\) when \(r_c\) is small, because \(F_c\) has eigenvalues at most 1. Consequently, we expect \(\mathrm{Tr}(\Sigma_{\mathrm{RL}}) > \mathrm{Tr}(\Sigma_{\mathrm{OPD}})\) in practice, implying that OPD follows a smoother and lower-noise optimization trajectory.

\paragraph{Summary.}
In the local regime, OPD can be approximated by a possibly degenerate convex quadratic minimization:
\begin{equation}
\min_{\Delta\theta}
\frac{1}{2}\Delta\theta^{\top}A\Delta\theta
-
b^{\top}\Delta\theta .
\end{equation}
The corresponding gradient descent dynamics admit the spectral form:
\begin{equation}
\Delta\theta_s
=
\sum_{i:\lambda_i>0}
\frac{1-(1-\eta\lambda_i)^s}{\lambda_i}
\beta_i u_i .
\end{equation}
This expression shows that the update along each eigen-direction is determined by the residual projection \(\beta_i=\langle b,u_i\rangle\), the local curvature \(\lambda_i\), and the finite-step growth factor \(1-(1-\eta\lambda_i)^s\).

If the driving term \(b\) is concentrated in a low-dimensional subspace, such as the top-\(k\) eigenspace of \(A\), and a clear spectral gap exists, then the update remains approximately confined to this subspace from the early stages of training. This provides a local explanation for \textbf{Early Low-Rank Lock-in}. 
At the module level, if a module has negligible coupling with the teacher residual, i.e., \(b_m\approx 0\), then its update is expected to be suppressed when cross-module coupling terms are not dominant. This explains \textbf{Functional Redundancy Avoidance}. 
Compared with RL, OPD benefits from a denser, lower-variance, and more directionally concentrated gradient signal, which helps explain the more concentrated and efficient update patterns observed in OPD.

\clearpage

\begin{tcolorbox}[title=Question, breakable]
\label{resoning chains}
Let $\triangle ABC$ be a triangle with $D$ on $\overline{BC}$ such that $\overline{AD}$ bisects $\angle BAC.$ Let $\omega$ be the circle that passes through $A$ and is tangent to segment $\overline{BC}$ at $D.$ Let $E \neq A$ and $F \neq A$ be the intersections of $\omega$ with segments $\overline{AB}$ and $\overline{AC},$ respectively. Suppose that $AB = 200, AC = 225,$ and all of $AE, AF, BD,$ and $CD$ are positive integers. Find the sum of all possible values of $BC.$
\end{tcolorbox}

\begin{tcolorbox}[title=Base Model, breakable]
We are given that Patrick, Tanya, and Jose start at different times and all                                                                                                    
arrive at the park at the same time. We need to find the distance $D = \dfrac{m}{n}$                                                                                           
in lowest terms, then compute $m+n$.                                                                                                                                           

\medskip                                                                                                                                                                       
\textbf{Step 1: Define Variables.}

Let $v$ be Patrick's speed (mph). Then Tanya's speed is $v+2$ and Jose's speed
is $v+9$. Let $t$ be Patrick's travel time (hours). Since they all arrive
simultaneously, Tanya travels $t-1$ hours and Jose travels $t-2$ hours, giving:
\[
D = vt = (v+2)(t-1) = (v+9)(t-2).
\]

\medskip
\textbf{Step 2: Solve for $t$ and $v$.}

From $vt = (v+2)(t-1)$:
\[
vt = vt - v + 2t - 2 \implies v = 2t-2. \tag{A}
\]

From $(v+2)(t-1) = (v+9)(t-2)$:
\[
vt - v + 2t - 2 = vt - 2v + 9t - 18 \implies v = 7t - 16. \tag{B}
\]

Setting (A) equal to (B):
\[
2t - 2 = 7t - 16 \implies 5t = 14 \implies t = \frac{14}{5}.
\]

Substituting back into (A):
\[
v = 2\cdot\frac{14}{5} - 2 = \frac{28}{5} - \frac{10}{5} = \frac{18}{5} \text{ mph}.
\]

\medskip
\textbf{Step 3: Compute the Distance.}

\[
D = vt = \frac{18}{5} \times \frac{14}{5} = \frac{252}{25}.
\]

Since $252 = 2^2\times 3^2\times 7$ and $25 = 5^2$ share no common factors,                                                                                                    
$\gcd(252,25)=1$, so the fraction is already in lowest terms.
      
\medskip        
\textbf{Final Answer.}
\[
m + n = 252 + 25 = \boxed{277}.
\]
\end{tcolorbox}

\begin{tcolorbox}[title=The early training stage model (unscaled), breakable]

We are given that Patrick starts walking at a constant rate, Tanya starts 1 hour after Patrick, and Jose starts 1 hour after Tanya. All three arrive at the park at the same   
time. Tanya runs 2 mph faster than Patrick walks, and Jose bicycles 7 mph faster than Tanya runs. The distance from school to park is $\tfrac{m}{n}$ in lowest terms; find
$m+n$.                                                                                                                                                                         
              
\medskip
\textbf{Step 1: Define Variables.}

Let $p$ be Patrick's walking speed (mph), so Tanya's running speed is $p+2$ and Jose's bicycling speed is $p+9$. Let $D$ be the distance (miles) and $T$ the total travel time
(hours) for Patrick.

\medskip
\textbf{Step 2: Set Up Equations.}

Since all three arrive simultaneously:
\begin{align}
D &= p \cdot T, \tag{1}\\
D &= (p+2)(T-1), \tag{2}\\
D &= (p+9)(T-2). \tag{3}
\end{align}

\medskip
\textbf{Step 3: Solve for $p$ and $T$.}

Equating (1) and (2):
\[
pT = (p+2)(T-1) = pT - p + 2T - 2
\implies p = 2T - 2. \tag{4}
\]

Equating (2) and (3):
\[
(p+2)(T-1) = (p+9)(T-2).
\]
Expanding both sides and simplifying:
\[
-p + 2T - 2 = -2p + 9T - 18
\implies p = 7T - 16. \tag{5}
\]

Setting (4) equal to (5):
\[
2T - 2 = 7T - 16 \implies 5T = 14 \implies T = \frac{14}{5}.
\]                                                                                                                                                                             

Substituting back into (4):                                                                                                                                                    
\[              
p = 2\cdot\frac{14}{5} - 2 = \frac{18}{5} \text{ mph}.
\]

\medskip
\textbf{Step 4: Compute the Distance.}

\[
D = pT = \frac{18}{5} \cdot \frac{14}{5} = \frac{252}{25}.
\]

Since $252 = 2^2 \cdot 3^2 \cdot 7$ and $25 = 5^2$ share no common factors, $\gcd(252,25)=1$.

\medskip
\textbf{Verification.}

\[
p = \frac{18}{5},\quad p+2 = \frac{28}{5},\quad p+9 = \frac{63}{5},
\]
\[
D = \frac{18}{5}\cdot\frac{14}{5} = \frac{252}{25},\quad
\frac{252/25}{28/5} = \frac{252}{25}\cdot\frac{5}{28} = \frac{9}{5} = T-1\ ,\quad
\frac{252/25}{63/5} = \frac{252}{25}\cdot\frac{5}{63} = \frac{4}{5} = T-2\ .
\]

\medskip
\textbf{Answer.}

\[
m + n = 252 + 25 = \boxed{277}.
\]

\end{tcolorbox}

\begin{tcolorbox}[title=The early training stage model (scaled):, breakable]
We are given a problem involving three people---Patrick, Tanya, and Jose---each
starting at different times and moving along the same path from school to the
park. All three arrive at the park at the same time. We need to find the distance from school to park, expressed as a reduced fraction $\dfrac{m}{n}$,
and then compute $m+n$.              
\medskip
\textbf{Step 1: Define Variables.}

Let $v$ be Patrick's walking speed (in miles per hour). Then:
\begin{itemize}
  \item Tanya's running speed is $v+2$ mph (2 mph faster than Patrick),
  \item Jose's bicycling speed is $(v+2)+7 = v+9$ mph (7 mph faster than Tanya).
\end{itemize}
Let $D$ be the distance from school to park (in miles), and let $t$ be the
time (in hours) that Patrick takes to walk from school to park. Since all three
arrive at the same time:
\begin{itemize}
  \item Patrick travels for $t$ hours,
  \item Tanya starts 1 hour after Patrick, so she travels for $t-1$ hours,
  \item Jose starts 1 hour after Tanya (2 hours after Patrick), so he travels
        for $t-2$ hours.
\end{itemize}
This gives the following time equations:
\[
t = \frac{D}{v}, \qquad
\frac{D}{v+2} = t-1, \qquad
\frac{D}{v+9} = t-2.
\]

\medskip                                                                                                                                                                      
\textbf{Step 2: Substitute $t = D/v$ into the Remaining Equations.}
 
Substituting $t = \dfrac{D}{v}$ into the equation for Tanya:
\[
\frac{D}{v+2} = \frac{D}{v} - 1.
\]
Substituting $t = \dfrac{D}{v}$ into the equation for Jose:
\[
\frac{D}{v+9} = \frac{D}{v} - 2.
\]
We now have two equations in two unknowns $D$ and $v$.

\medskip
\textbf{Step 3: Derive Equation (A) from Tanya's Equation.}

Starting from:
\[
\frac{D}{v+2} = \frac{D}{v} - 1.
\]
Multiply both sides by $v(v+2)$:
\[
D \cdot v = D(v+2) - v(v+2).
\]
Expand the right-hand side:
\[
Dv = Dv + 2D - v^2 - 2v.
\]
Subtract $Dv$ from both sides:
\[
0 = 2D - v^2 - 2v.
\]
Rearranging:
\[
v^2 + 2v - 2D = 0. \tag{A}
\]

\medskip
\textbf{Step 4: Derive Equation (B) from Jose's Equation.}

Starting from:
\[
\frac{D}{v+9} = \frac{D}{v} - 2.
\]
Multiply both sides by $v(v+9)$:
\[
D \cdot v = D(v+9) - 2v(v+9).
\]
Expand the right-hand side:
\[
Dv = Dv + 9D - 2v^2 - 18v.
\]
Subtract $Dv$ from both sides:
\[
0 = 9D - 2v^2 - 18v.
\]
Rearranging:
\[
2v^2 + 18v - 9D = 0. \tag{B}
\]

\medskip
\textbf{Step 5: Solve the System of Equations (A) and (B).}

From equation (A):
\[
v^2 + 2v - 2D = 0 \implies D = \frac{v^2 + 2v}{2}.
\]
Substitute this expression for $D$ into equation (B):
\[
2v^2 + 18v - 9 \cdot \frac{v^2+2v}{2} = 0.
\]
Multiply every term by 2 to clear the denominator:
\[
4v^2 + 36v - 9(v^2+2v) = 0.
\]
Expand:
\[
4v^2 + 36v - 9v^2 - 18v = 0.
\]
Combine like terms:
\[
(4v^2 - 9v^2) + (36v - 18v) = 0 \implies -5v^2 + 18v = 0.
\]
Factor:
\[
v(-5v + 18) = 0.
\]
This gives $v = 0$ or $v = \dfrac{18}{5}$. Since Patrick is walking, $v > 0$,
so:
\[
v = \frac{18}{5} \text{ mph}.
\]

\medskip
\textbf{Step 6: Compute the Distance $D$.}

Using $D = \dfrac{v^2+2v}{2}$ with $v = \dfrac{18}{5}$:
\[
v^2 = \left(\frac{18}{5}\right)^2 = \frac{324}{25},
\qquad
2v = 2 \times \frac{18}{5} = \frac{36}{5} = \frac{180}{25}.
\]
\[
v^2 + 2v = \frac{324}{25} + \frac{180}{25} = \frac{504}{25}.
\]
\[
D = \frac{504/25}{2} = \frac{504}{50} = \frac{252}{25}.
\]

\medskip
\textbf{Step 7: Verify with All Three Travelers.}

Patrick's travel time: $t = \dfrac{D}{v} = \dfrac{252/25}{18/5}
= \dfrac{252}{25} \times \dfrac{5}{18} = \dfrac{1260}{450} = \dfrac{14}{5}$
hours.

\medskip
\textit{Tanya} travels for $t - 1 = \dfrac{14}{5} - \dfrac{5}{5} = \dfrac{9}{5}$
hours at speed $v+2 = \dfrac{18}{5}+\dfrac{10}{5} = \dfrac{28}{5}$ mph:
\[
\frac{28}{5} \times \frac{9}{5} = \frac{252}{25} = D. \quad \checkmark
\]

\textit{Jose} travels for $t - 2 = \dfrac{14}{5} - \dfrac{10}{5} = \dfrac{4}{5}$
hours at speed $v+9 = \dfrac{18}{5}+\dfrac{45}{5} = \dfrac{63}{5}$ mph:
\[
\frac{63}{5} \times \frac{4}{5} = \frac{252}{25} = D. \quad \checkmark
\]
All three cover the same distance and arrive at the same time.

\medskip
\textbf{Step 8: Confirm the Fraction is in Lowest Terms.}

Factoring numerator and denominator:
\[
252 = 2^2 \times 3^2 \times 7, \qquad 25 = 5^2.
\]
Since 252 contains no factor of 5 (as $252 \div 5 = 50.4$ is not an integer)
and 25 contains no factor of 2, 3, or 7, we have $\gcd(252, 25) = 1$.
Therefore $\dfrac{252}{25}$ is already in lowest terms, giving $m = 252$ and
$n = 25$.

\medskip
\textbf{Final Answer.}
\[
m + n = 252 + 25 = \boxed{277}.
\]

\end{tcolorbox}
\begin{tcolorbox}[title=Teacher Model, breakable]
We are given a problem involving three people---Patrick, Tanya, and Jose---each                                                                                                
starting at different times and moving along the same road from school to the                                                                                                  
park. They all arrive at the park at the same time. We need to find the distance                                                                                               
from school to park, expressed as a reduced fraction $\dfrac{m}{n}$, and then
compute $m+n$.                                                                                                                                                                 
              
\medskip                                                                                                                                                                       
\textbf{Step 1: Define Variables.}

Let $d$ be the distance from school to park (in miles) and $v$ be Patrick's
walking speed (in mph). Then:
\begin{itemize}
  \item Tanya's running speed is $v+2$ mph (2 mph faster than Patrick),
  \item Jose's bicycling speed is $(v+2)+7 = v+9$ mph (7 mph faster than Tanya).
\end{itemize}

\medskip
\textbf{Step 2: Determine Travel Times.}

Let $t$ be the number of hours Patrick takes to reach the park. Since Tanya
starts 1 hour after Patrick and arrives at the same time, she travels for
$t-1$ hours. Since Jose starts 1 hour after Tanya (i.e., 2 hours after
Patrick) and arrives at the same time, he travels for $t-2$ hours.

Using distance $=$ speed $\times$ time, we obtain three equations:
\begin{align}
v &= \frac{d}{t}, \tag{1}\\[6pt]
v+2 &= \frac{d}{t-1}, \tag{2}\\[6pt]
v+9 &= \frac{d}{t-2}. \tag{3}
\end{align}

\medskip
\textbf{Step 3: Derive Equation (A) from Equations (1) and (2).}

From equation (1): $t = \dfrac{d}{v}$. Substituting into equation (2):
\[
v+2 = \frac{d}{\dfrac{d}{v}-1} = \frac{d}{\dfrac{d-v}{v}} = \frac{dv}{d-v}.
\]
Multiply both sides by $(d-v)$:
\[
(v+2)(d-v) = dv.
\]
Expand the left-hand side:
\[
vd - v^2 + 2d - 2v = dv.
\]
Subtract $dv$ from both sides:
\[
-v^2 + 2d - 2v = 0.
\]
Rearranging:
\[
v^2 + 2v - 2d = 0. \tag{A}
\]

\medskip
\textbf{Step 4: Derive Equation (B) from Equations (1) and (3).}

Again using $t = \dfrac{d}{v}$, we have $t - 2 = \dfrac{d}{v} - 2 = \dfrac{d-2v}{v}$.
Substituting into equation (3):
\[
v+9 = \frac{d}{\dfrac{d-2v}{v}} = \frac{dv}{d-2v}.
\]
Multiply both sides by $(d-2v)$:
\[
(v+9)(d-2v) = dv.
\]
Expand the left-hand side:
\[
vd - 2v^2 + 9d - 18v = dv.
\]
Subtract $dv$ from both sides:
\[
-2v^2 + 9d - 18v = 0.
\]
Rearranging:
\[
2v^2 + 18v - 9d = 0. \tag{B}
\]

\medskip
\textbf{Step 5: Solve the System of Equations (A) and (B).}

From equation (A):
\[
2d = v^2 + 2v \implies d = \frac{v^2+2v}{2}.
\]
Substitute this expression for $d$ into equation (B):
\[
2v^2 + 18v - 9\cdot\frac{v^2+2v}{2} = 0.
\]
Multiply every term by 2 to clear the denominator:
\[
4v^2 + 36v - 9(v^2+2v) = 0.
\]
Expand:
\[
4v^2 + 36v - 9v^2 - 18v = 0.
\]
Combine like terms:
\[
(4v^2 - 9v^2) + (36v - 18v) = 0 \implies -5v^2 + 18v = 0.
\]
Factor:
\[
v(-5v + 18) = 0.
\]
This gives $v = 0$ or $v = \dfrac{18}{5}$. Since Patrick is walking, $v > 0$,
so:
\[
v = \frac{18}{5} \text{ mph}.
\]

\medskip
\textbf{Step 6: Compute the Speeds.}

\begin{itemize}
  \item Patrick's speed: $v = \dfrac{18}{5}$ mph,
  \item Tanya's speed: $v+2 = \dfrac{18}{5} + \dfrac{10}{5} = \dfrac{28}{5}$ mph,
  \item Jose's speed: $v+9 = \dfrac{18}{5} + \dfrac{45}{5} = \dfrac{63}{5}$ mph.
\end{itemize}

\medskip
\textbf{Step 7: Compute the Distance $d$.}

\[
d = \frac{v^2 + 2v}{2}.
\]
First compute $v^2$:
\[
v^2 = \left(\frac{18}{5}\right)^2 = \frac{324}{25}.
\]
Then compute $2v$:
\[
2v = 2 \times \frac{18}{5} = \frac{36}{5} = \frac{180}{25}.
\]
Add:
\[
v^2 + 2v = \frac{324}{25} + \frac{180}{25} = \frac{504}{25}.
\]
Divide by 2:
\[
d = \frac{504}{25} \div 2 = \frac{504}{50} = \frac{252}{25}.
\]

\medskip
\textbf{Step 8: Verify the Answer.}

Patrick's total travel time:
\[
t = \frac{d}{v} = \frac{252/25}{18/5} = \frac{252}{25} \times \frac{5}{18}
= \frac{252 \times 5}{25 \times 18} = \frac{1260}{450} = \frac{14}{5} \text{ hours}.
\]

Tanya travels for $t - 1 = \dfrac{14}{5} - \dfrac{5}{5} = \dfrac{9}{5}$ hours at
speed $\dfrac{28}{5}$ mph:
\[
\frac{28}{5} \times \frac{9}{5} = \frac{252}{25}\ \checkmark.
\]

Jose travels for $t - 2 = \dfrac{14}{5} - \dfrac{10}{5} = \dfrac{4}{5}$ hours at
speed $\dfrac{63}{5}$ mph:
\[
\frac{63}{5} \times \frac{4}{5} = \frac{252}{25}\ \checkmark.
\]

All three cover the same distance $\dfrac{252}{25}$ miles.

\medskip
\textbf{Step 9: Confirm the Fraction is in Lowest Terms.}

Factoring the numerator and denominator:
\[
252 = 2^2 \times 3^2 \times 7, \qquad 25 = 5^2.
\]
Since 252 and 25 share no common prime factors, $\gcd(252, 25) = 1$, so
$\dfrac{252}{25}$ is already in lowest terms.

\medskip
\textbf{Final Answer.}
\[
m + n = 252 + 25 = \boxed{277}.
\]

\label{example}
\end{tcolorbox}

\begin{figure}[h] 
    \centering
    \includegraphics[width=1\textwidth]{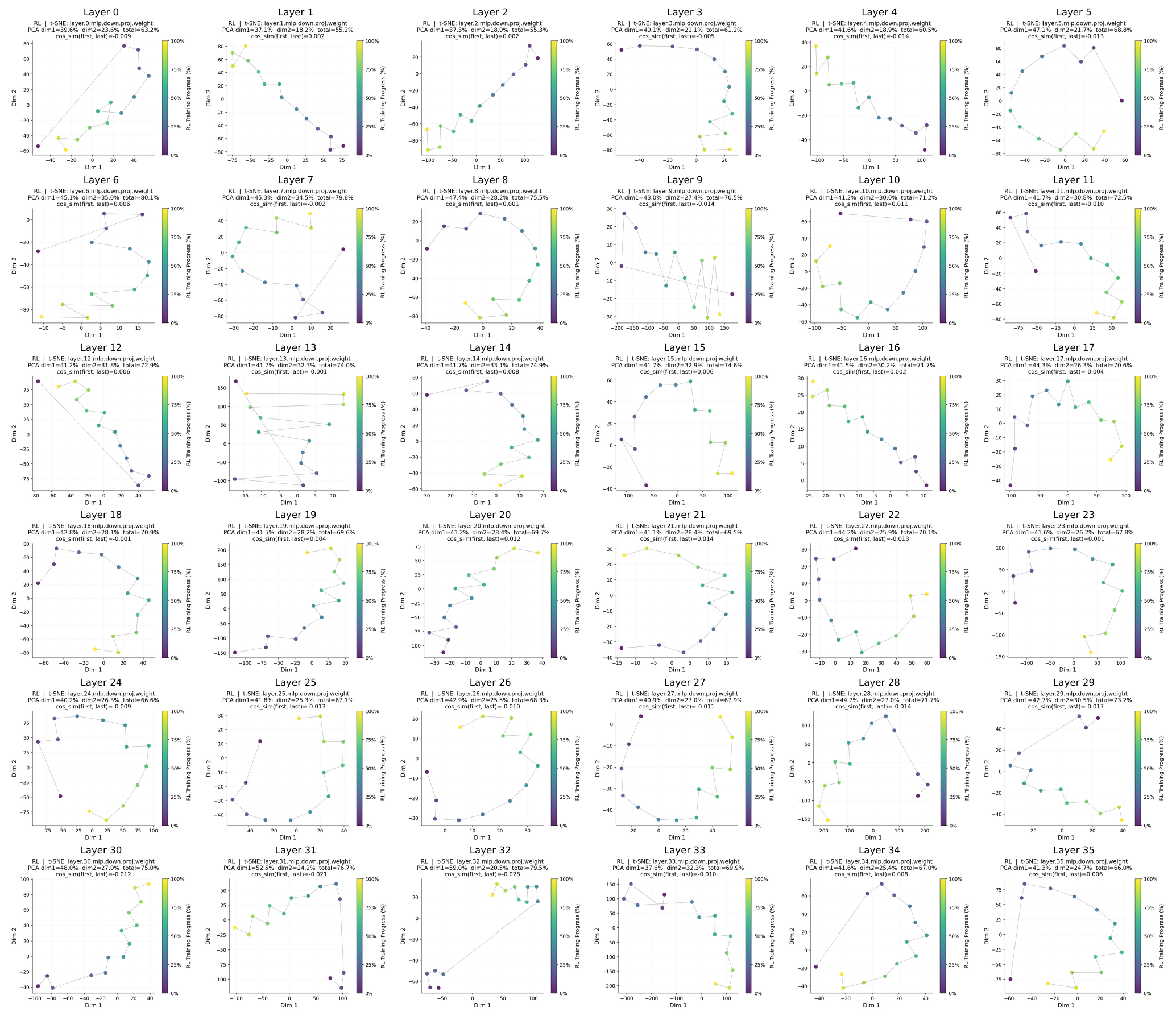} 
    \caption{t-SNE visualization of $\mathcal{U}_1$ trajectories under DAPO for MLP modules.}
    \label{tsne_grid_mlp_down_proj}
\end{figure}

\begin{figure}[h] 
    \centering
    \includegraphics[width=1\textwidth]{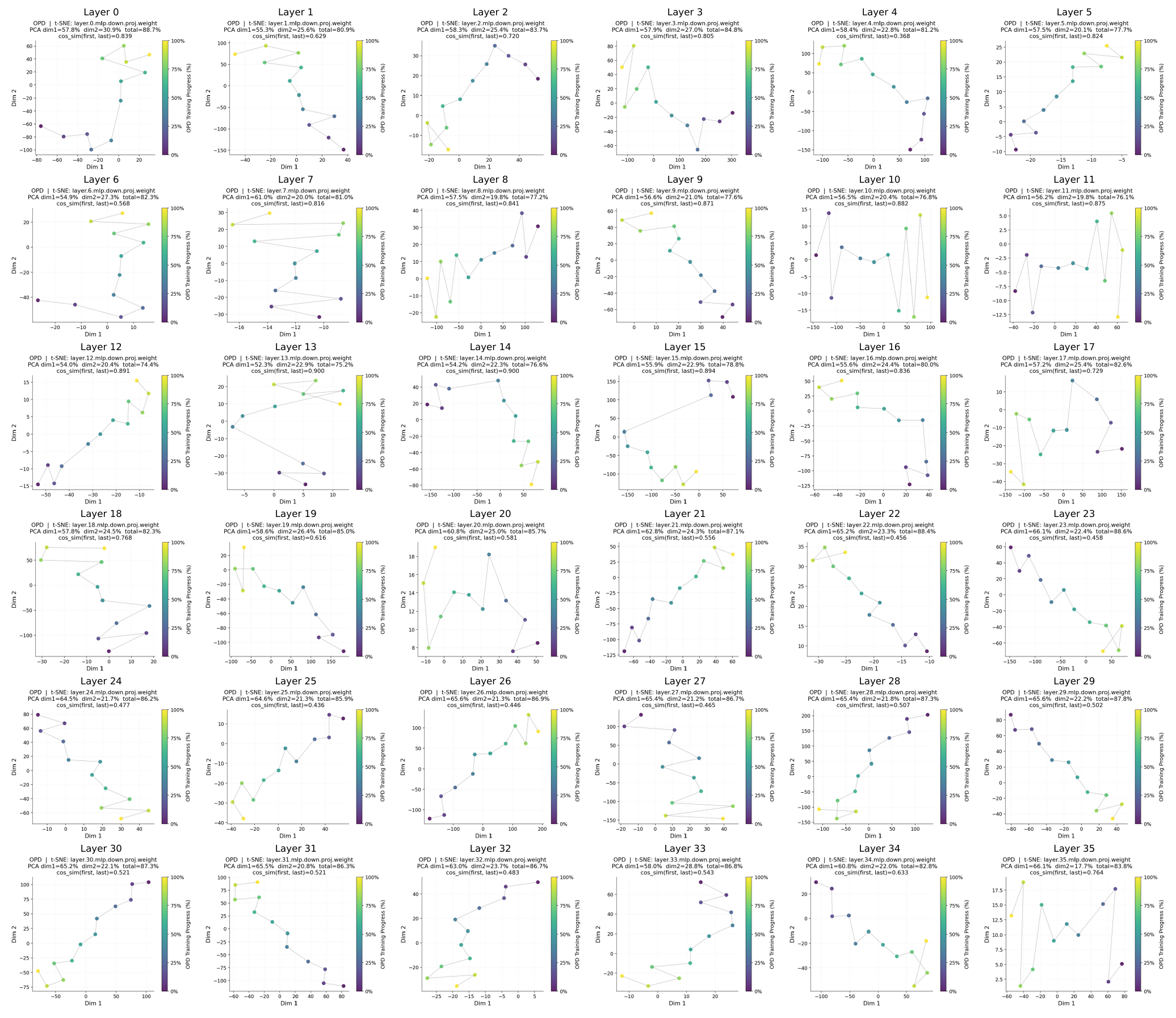} 
    \caption{t-SNE visualization of $\mathcal{U}_1$ trajectories under OPD for MLP modules.}
\end{figure}

\begin{figure}[h] 
    \centering
    \includegraphics[width=1\textwidth]{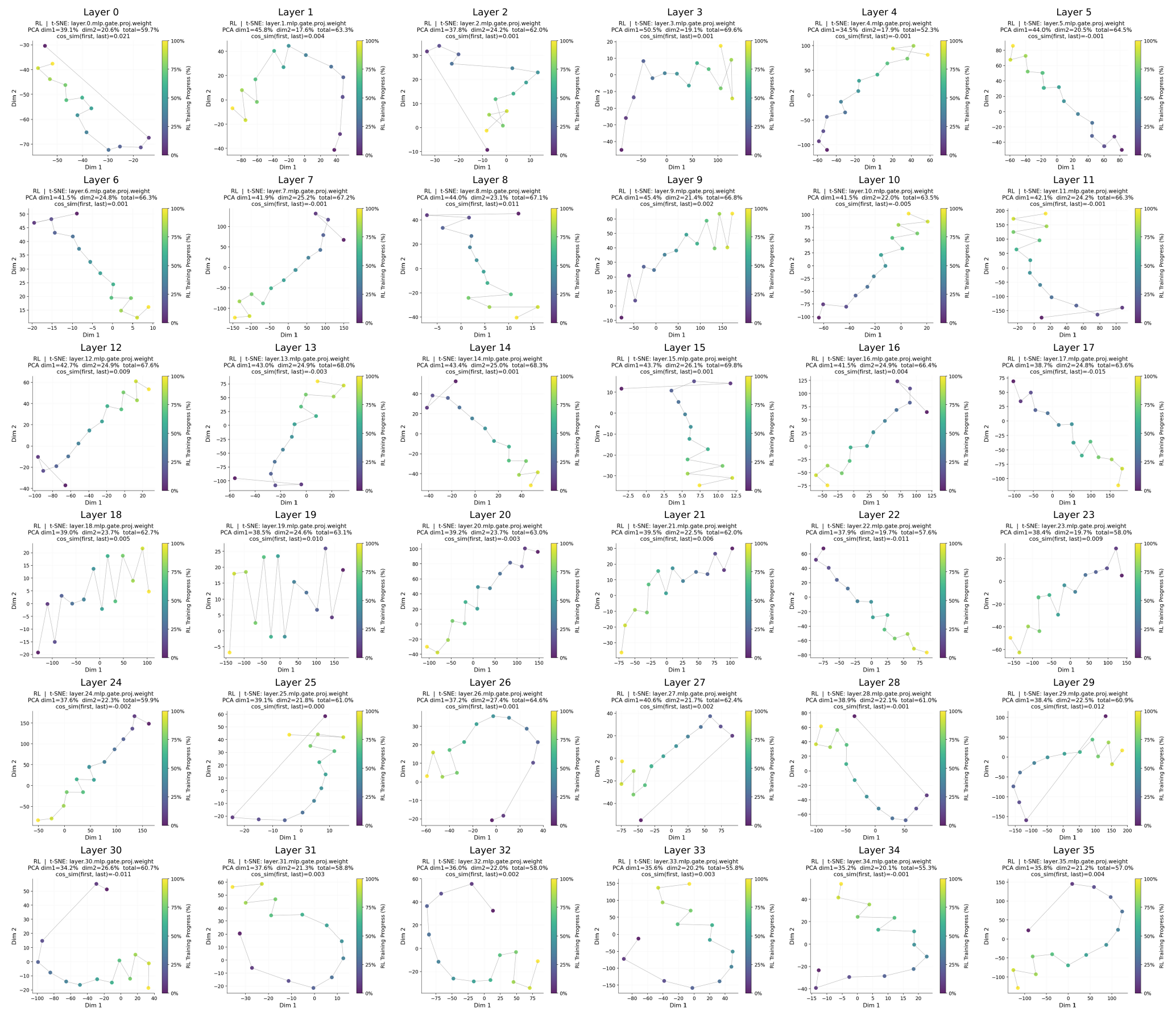} 
    \caption{t-SNE visualization of $\mathcal{U}_1$ trajectories under DAPO for MLP GATE modules.}
\end{figure}

\begin{figure}[h] 
    \centering
    \includegraphics[width=1\textwidth]{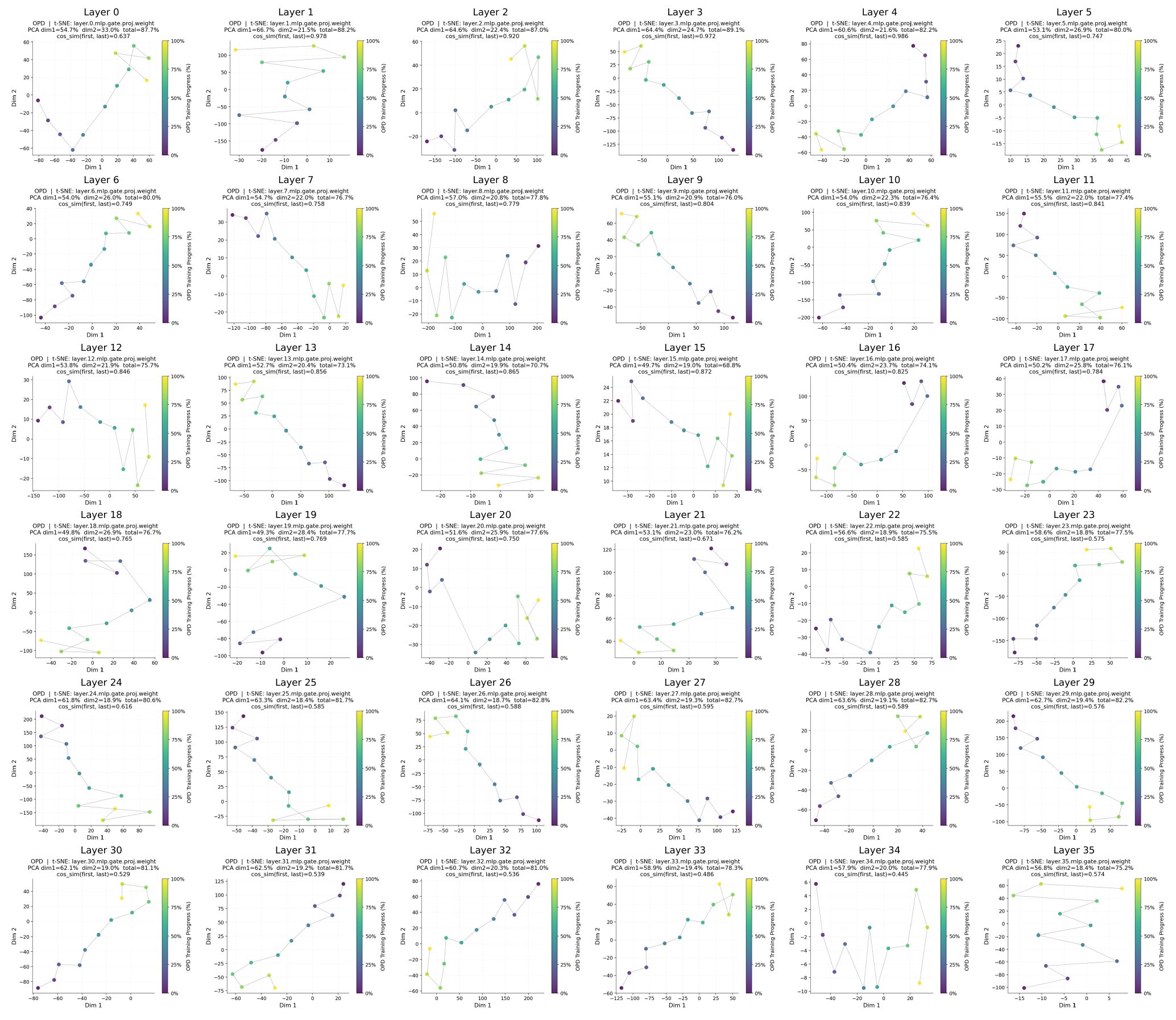} 
    \caption{t-SNE visualization of $\mathcal{U}_1$ trajectories under OPD for MLP GATE modules.}
\end{figure}

\begin{figure}[h] 
    \centering
    \includegraphics[width=1\textwidth]{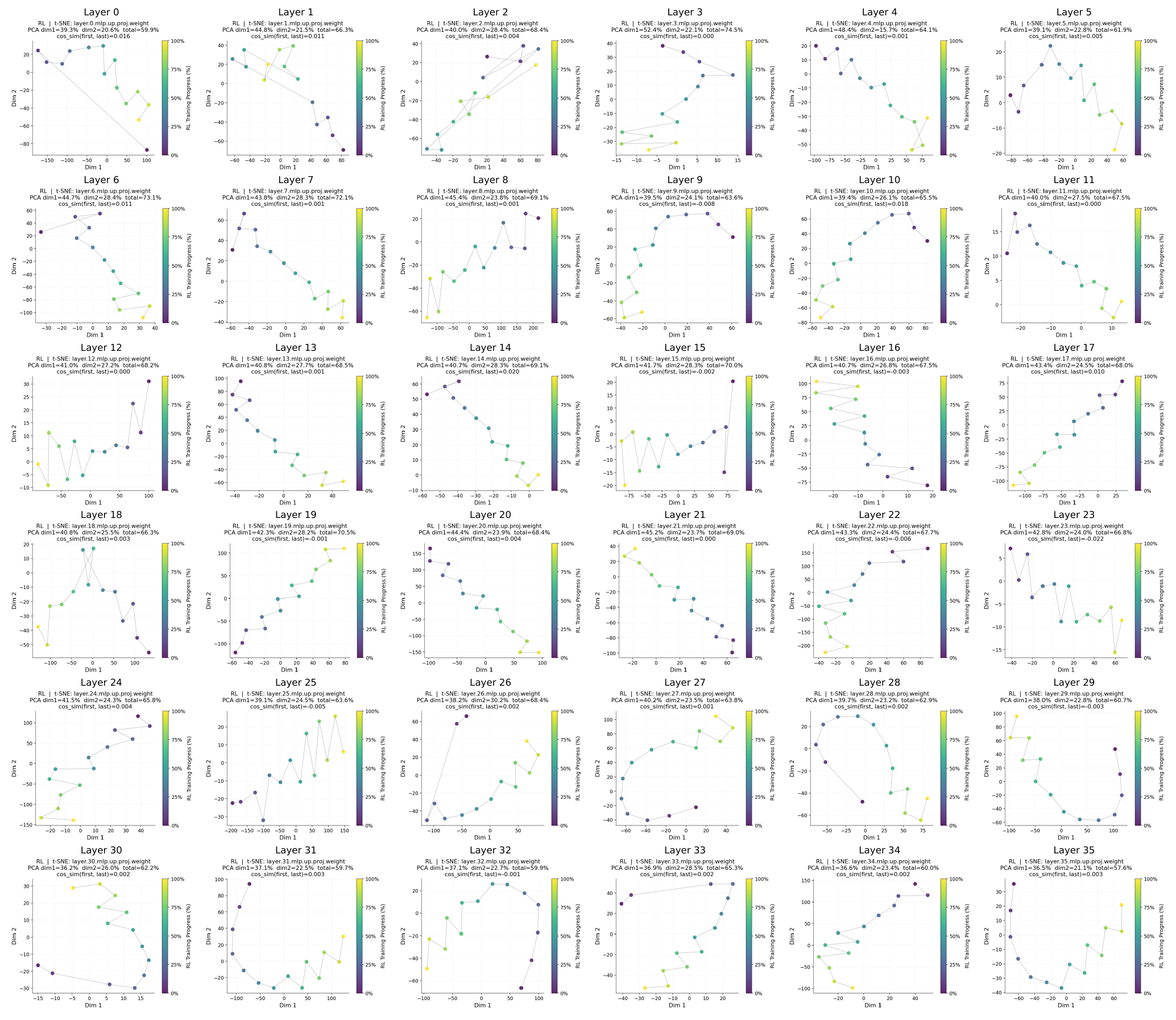} 
    \caption{t-SNE visualization of $\mathcal{U}_1$ trajectories under DAPO for MLP UP modules.}
\end{figure}

\begin{figure}[h] 
    \centering
    \includegraphics[width=1\textwidth]{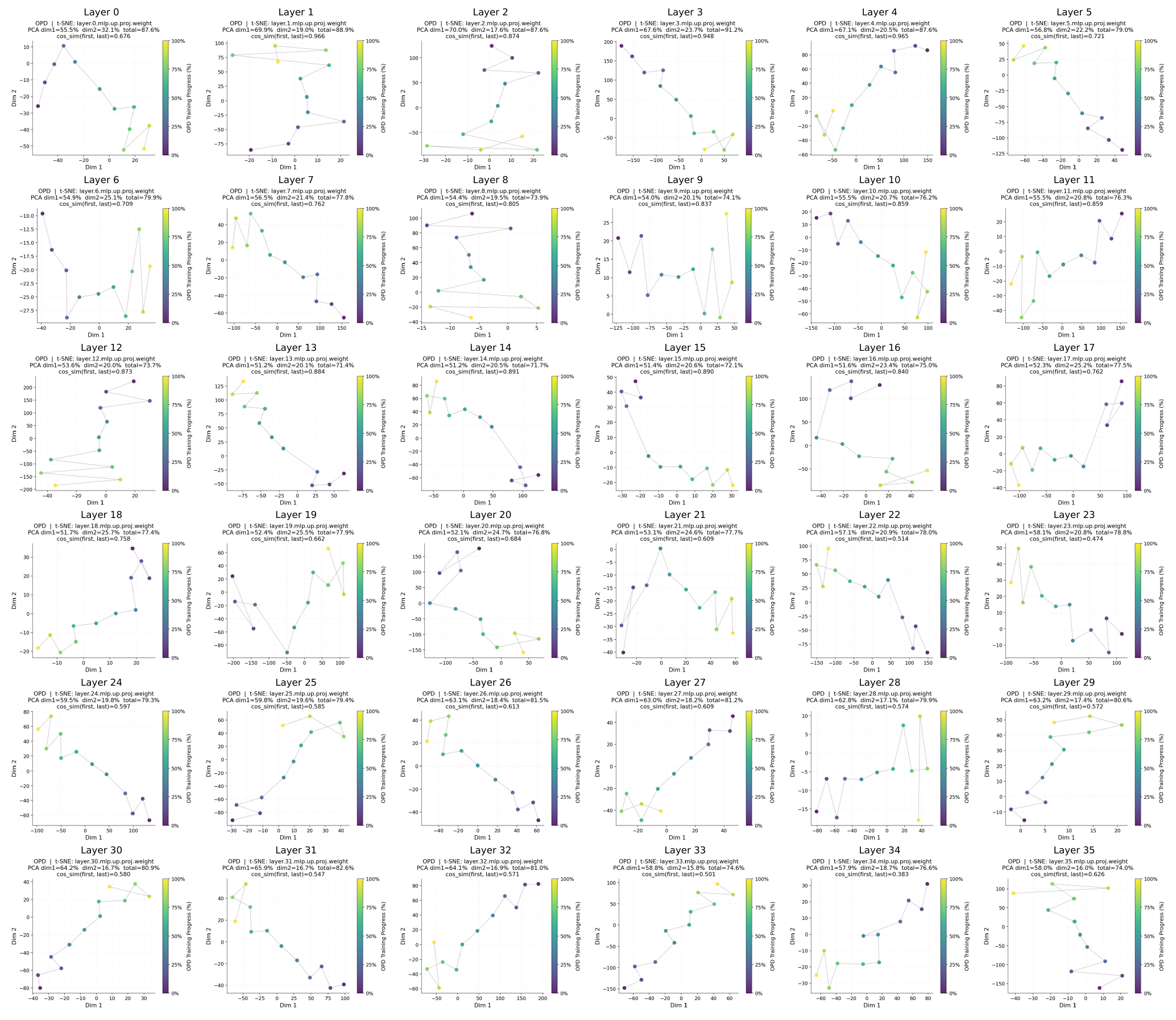} 
    \caption{t-SNE visualization of $\mathcal{U}_1$ trajectories under OPD for MLP UP modules.}
\end{figure}

\begin{figure}[h] 
    \centering
    \includegraphics[width=1\textwidth]{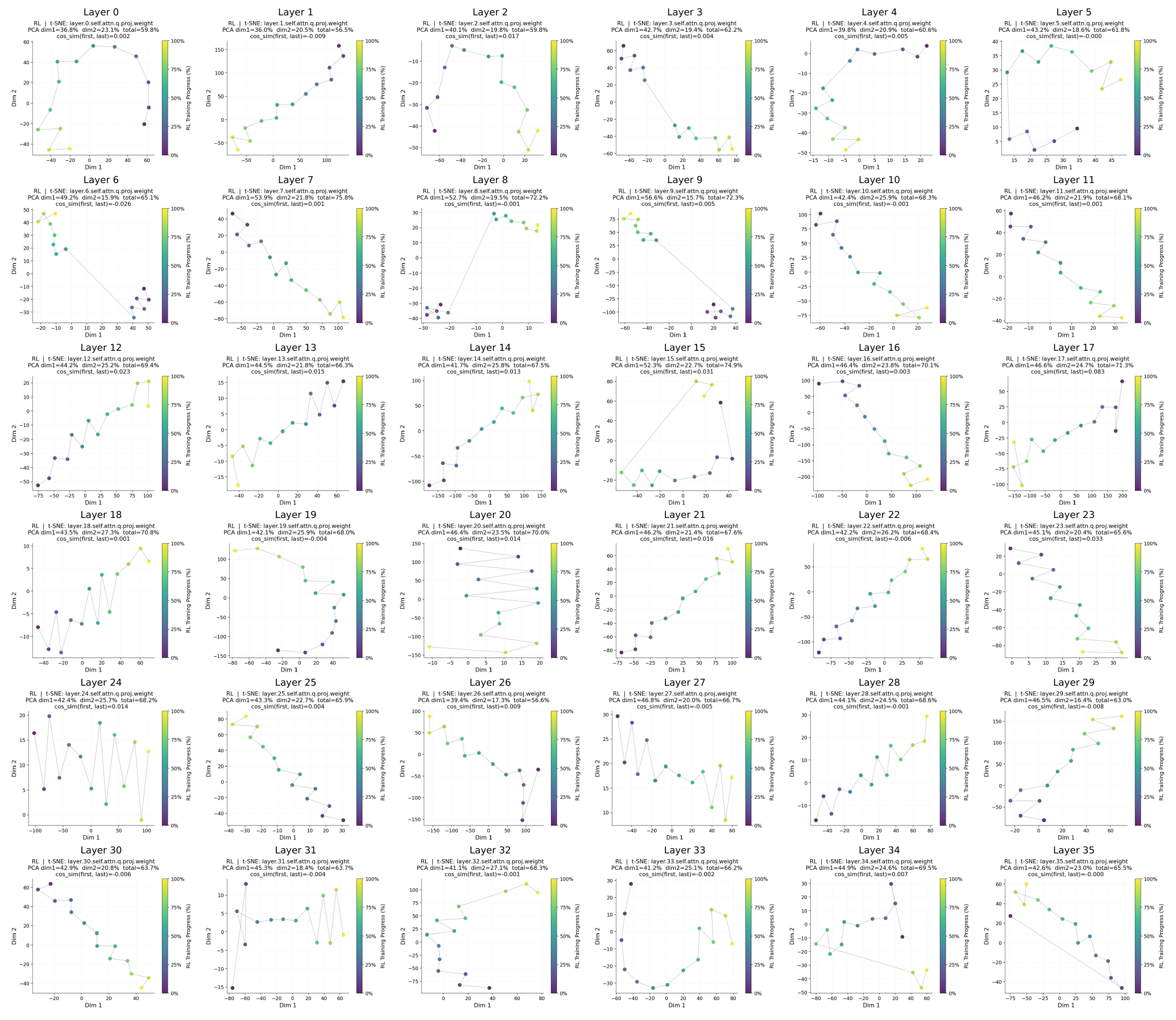} 
    \caption{t-SNE visualization of $\mathcal{U}_1$ trajectories under DAPO for Attn Q modules.}
\end{figure}

\begin{figure}[h] 
    \centering
    \includegraphics[width=1\textwidth]{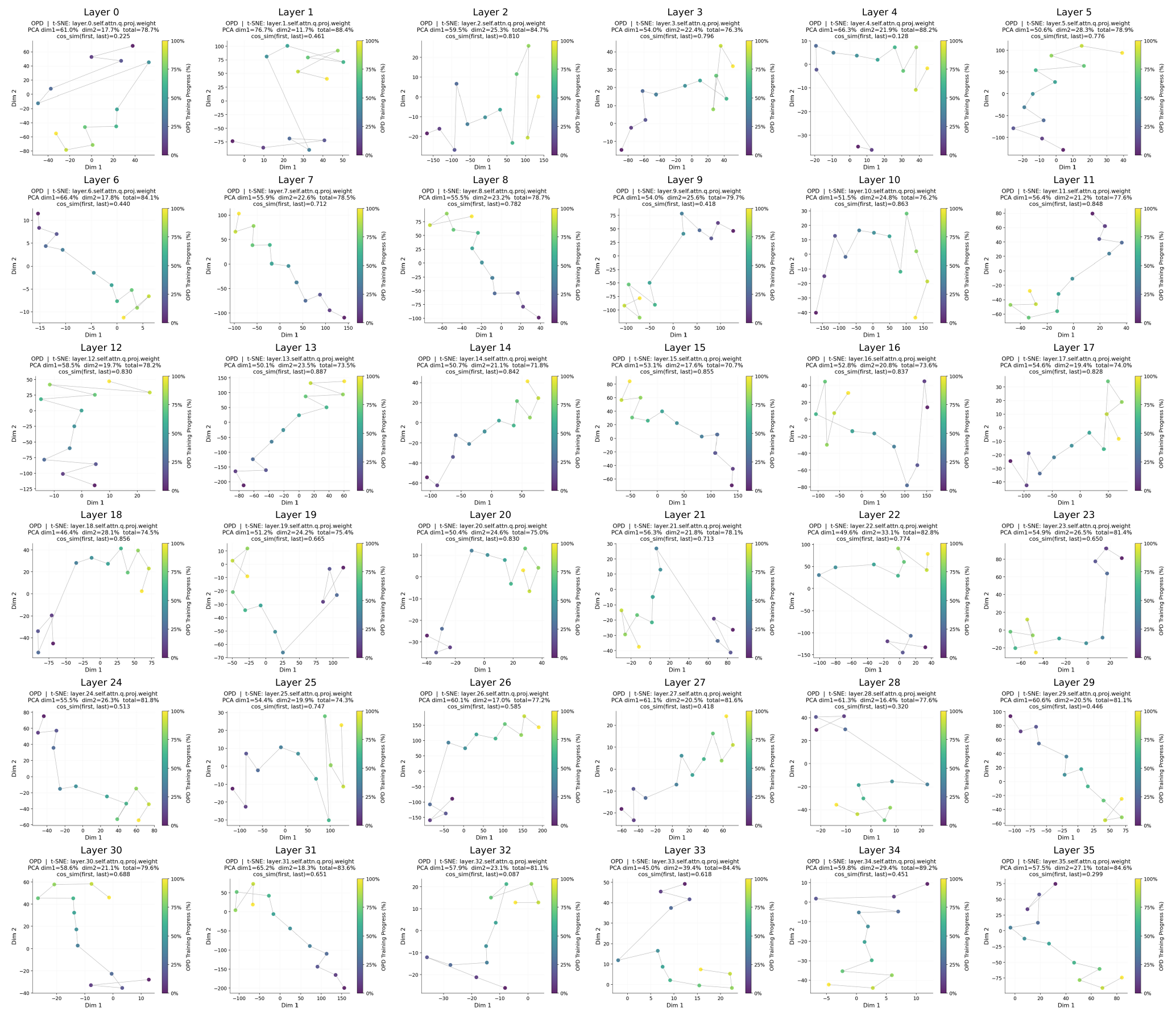} 
    \caption{t-SNE visualization of $\mathcal{U}_1$ trajectories under OPD for Attn Q modules.}
\end{figure}

\begin{figure}[h] 
    \centering
    \includegraphics[width=1\textwidth]{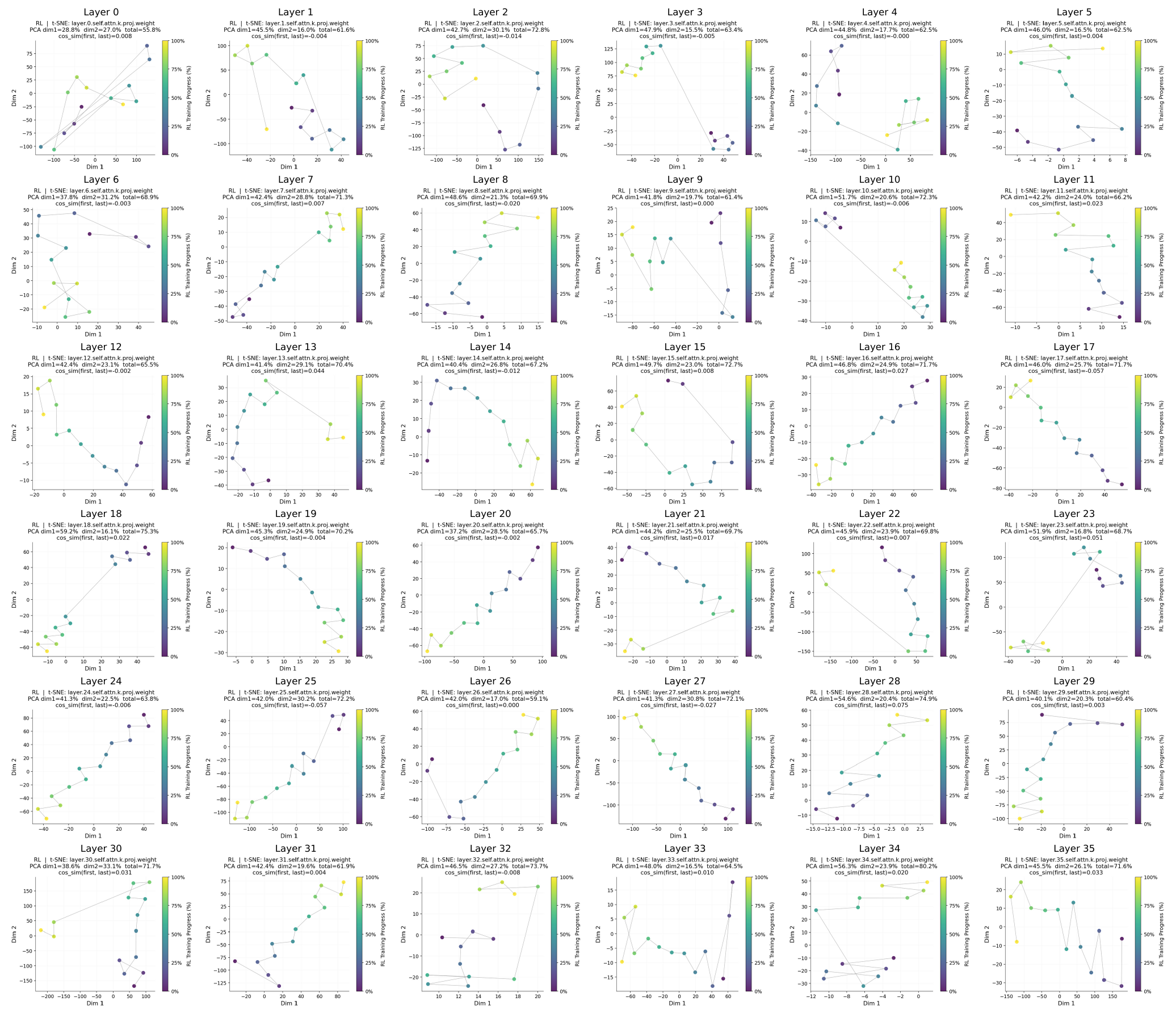} 
    \caption{t-SNE visualization of $\mathcal{U}_1$ trajectories under DAPO for Attn K modules.}
\end{figure}

\begin{figure}[h] 
    \centering
    \includegraphics[width=1\textwidth]{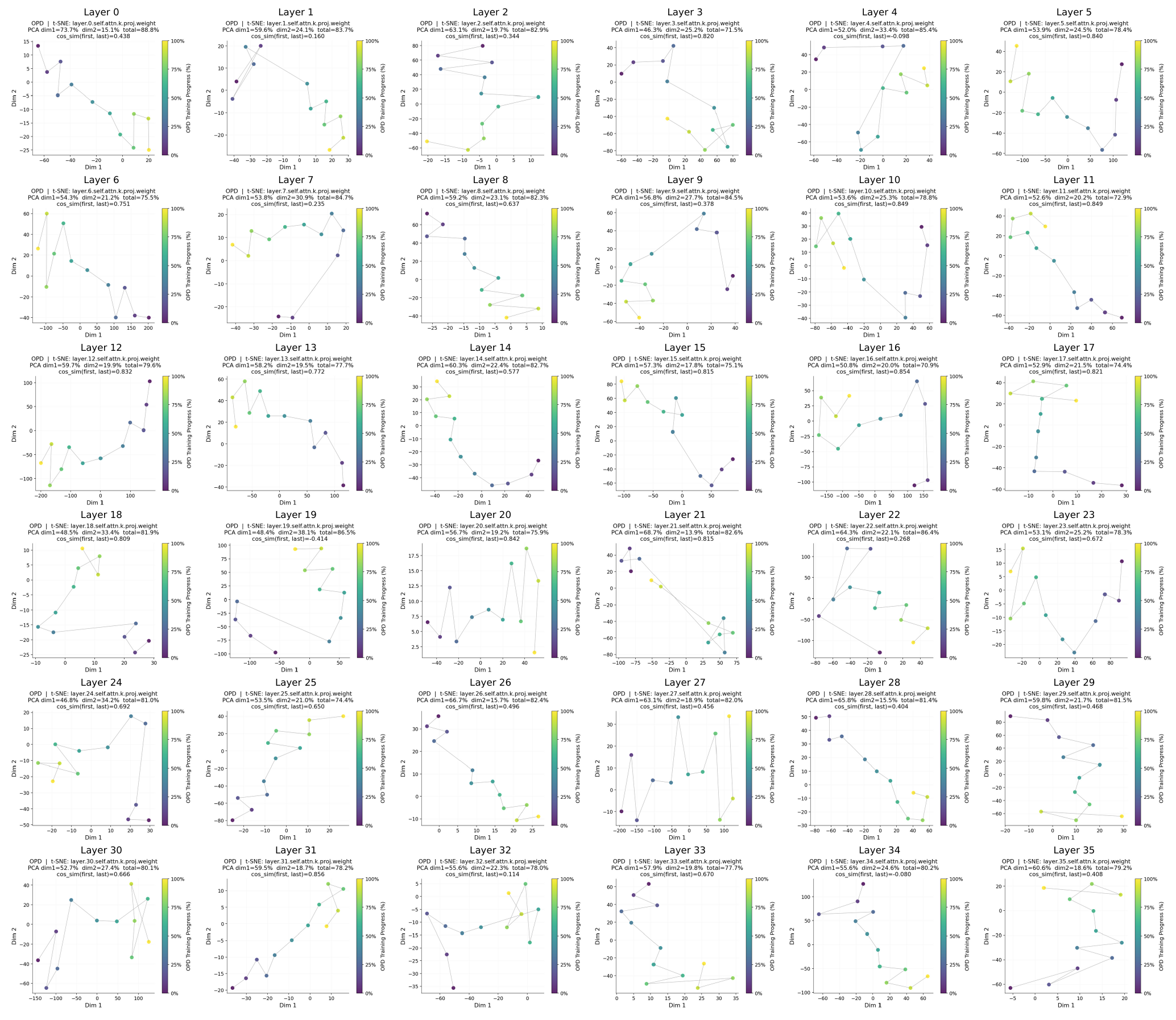} 
    \caption{t-SNE visualization of $\mathcal{U}_1$ trajectories under OPD for Attn K modules.}
\end{figure}

\begin{figure}[h] 
    \centering
    \includegraphics[width=1\textwidth]{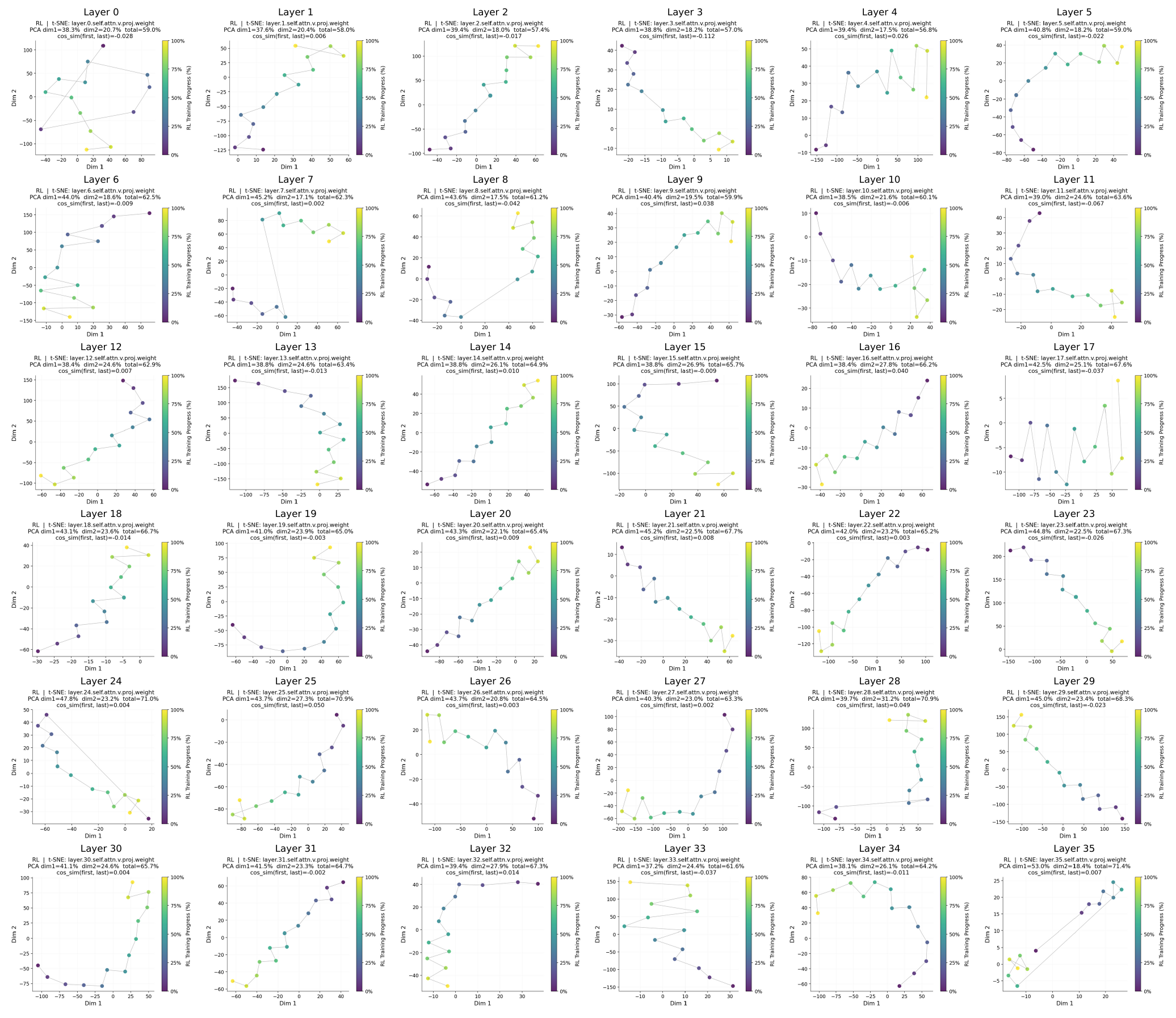} 
    \caption{t-SNE visualization of $\mathcal{U}_1$ trajectories under DAPO for Attn V modules.}
\end{figure}

\begin{figure}[h] 
    \centering
    \includegraphics[width=1\textwidth]{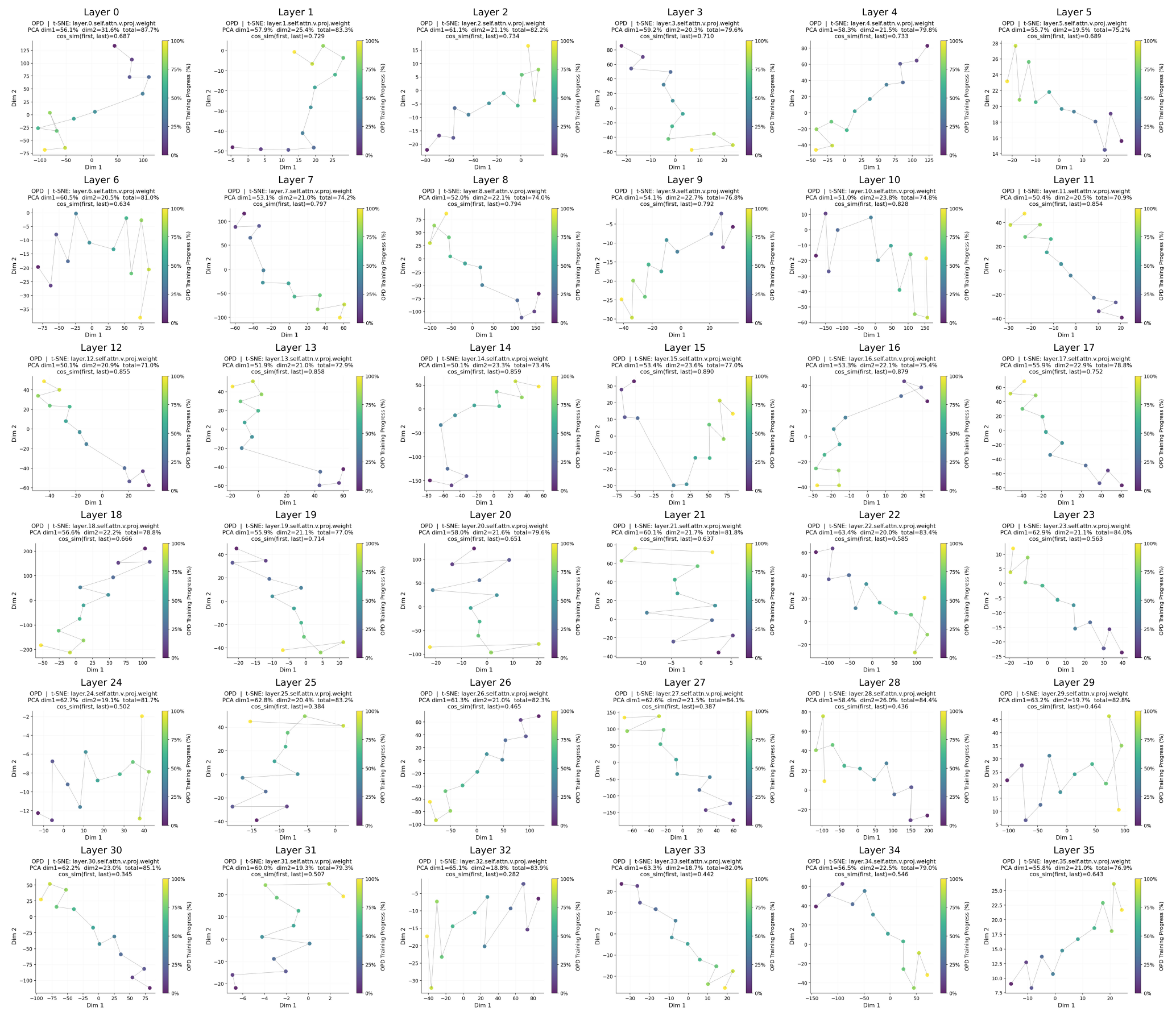} 
    \caption{t-SNE visualization of $\mathcal{U}_1$ trajectories under OPD for Attn V modules.}
\end{figure}

\begin{figure}[h] 
    \centering
    \includegraphics[width=1\textwidth]{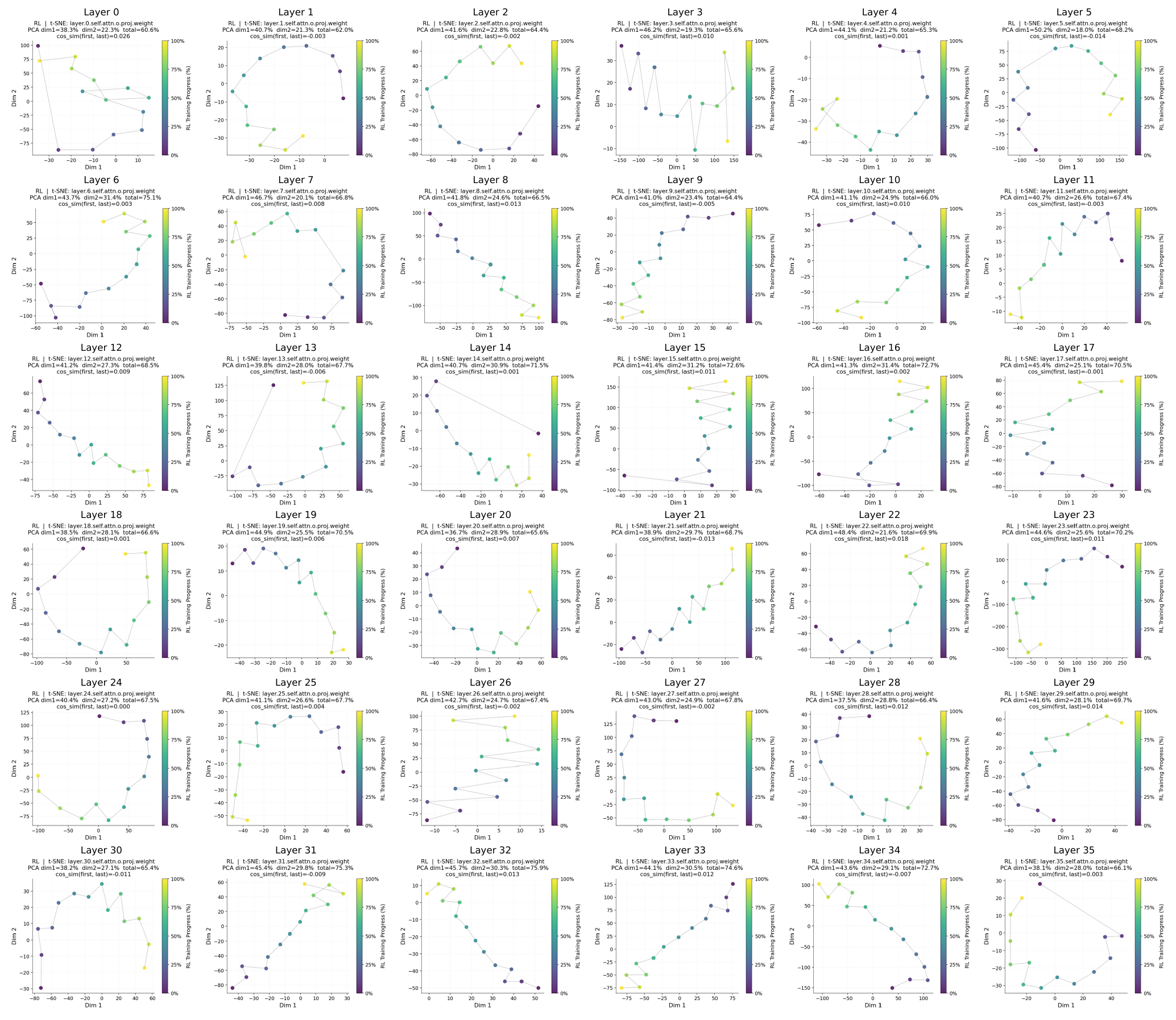} 
    \caption{t-SNE visualization of $\mathcal{U}_1$ trajectories under DAPO for Attn modules.}
\end{figure}

\begin{figure}[h] 
    
    \centering
    \includegraphics[width=1\textwidth]{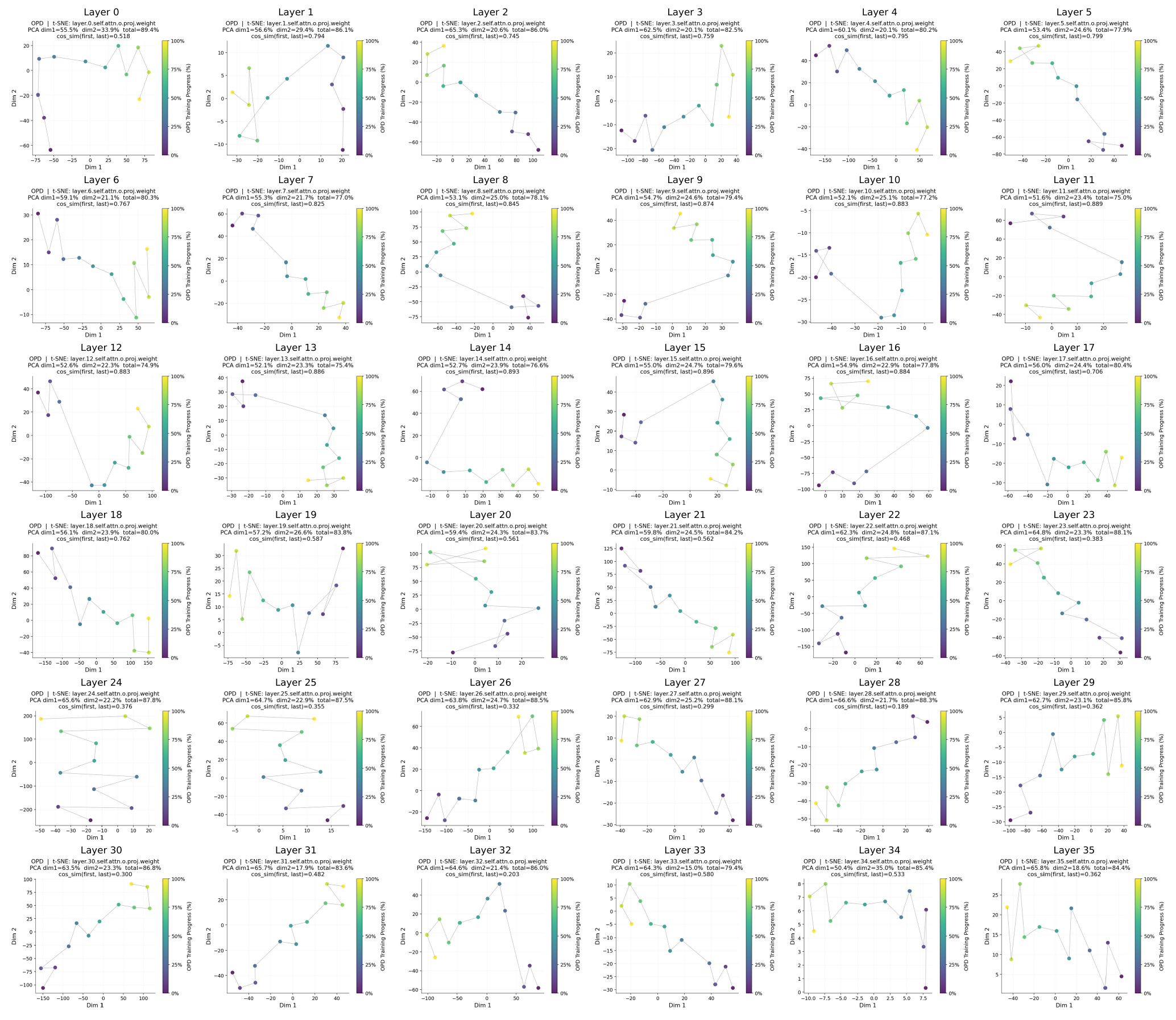} 
    \caption{t-SNE visualization of $\mathcal{U}_1$ trajectories under OPD for Attn modules.}
    \label{tsne_grid_self_attn_o_proj (1)}
\end{figure}

\end{document}